\documentclass[runningheads]{llncs}
\usepackage{graphicx}

\usepackage{tikz}
\usepackage{comment}
\usepackage{amsmath,amssymb} 
\usepackage{color}
\usepackage{subcaption}
\usepackage{xcolor,colortbl}

\definecolor{vvlightgray}{rgb}{0.9,0.9,0.9}
\definecolor{vlightgray}{rgb}{0.8,0.8,0.8}
\usepackage[accsupp]{axessibility}  
\usepackage{booktabs}
\usepackage[nokeyprefix]{refstyle}
\usepackage{varioref}
\usepackage{xr-hyper}
\usepackage[pagebackref,breaklinks,colorlinks]{hyperref}
\usepackage[ruled]{algorithm2e}

\usepackage{xcolor,colortbl}
\usepackage{multirow}
\newcommand{\red}[1]{\textcolor{red}{#1}}
\newcommand{\blue}[1]{\textcolor{blue}{#1}}

\begin{document}
\pagestyle{headings}
\mainmatter
\def\ECCVSubNumber{3991}  

\title{A Non-isotropic Probabilistic Take on Proxy-based Deep Metric Learning} 


\titlerunning{Non-isotropic Probabilistic Proxy-based DML}
%
\author{Michael Kirchhof\inst{1}$^{, *}$ \and
Karsten Roth\inst{1}$^{, *}$ \and
Zeynep Akata\inst{1} \and Enkelejda Kasneci\inst{1}}
\authorrunning{M. Kirchhof et al.}
%
\institute{$^1$University of Tübingen, Germany\\($^*$) equal contribution}
\maketitle

\begin{abstract}
Proxy-based Deep Metric Learning (DML) learns deep representations by embedding images close to their class representatives (\textit{proxies}), commonly with respect to the angle between them.
However, this disregards the embedding norm, which can carry additional beneficial context such as class- or image-intrinsic uncertainty. In addition, proxy-based DML struggles to learn class-internal structures.
To address both issues at once, we introduce non-isotropic probabilistic proxy-based DML. 
We model images as directional von Mises-Fisher (vMF) distributions on the hypersphere that can reflect image-intrinsic uncertainties. Further, we derive non-isotropic von Mises-Fisher (nivMF) distributions for class proxies to better represent complex class-specific variances. To measure the proxy-to-image distance between these models, we develop and investigate multiple distribution-to-point and distribution-to-distribution metrics. 
Each framework choice is motivated by a set of ablational studies, which showcase beneficial properties of our probabilistic approach to proxy-based DML, such as uncertainty-awareness, better behaved gradients during training, and overall improved generalization performance.
The latter is especially reflected in the competitive performance on the standard DML benchmarks, where our  approach compares favourably, suggesting that existing proxy-based DML can significantly benefit from a more probabilistic treatment. Code is available at \url{github.com/ExplainableML/Probabilistic_Deep_Metric_Learning}.
\end{abstract}

\begin{figure}
    \centering
    \includegraphics[width=0.85\linewidth, trim={1cm 2cm 1cm 2cm}, clip]{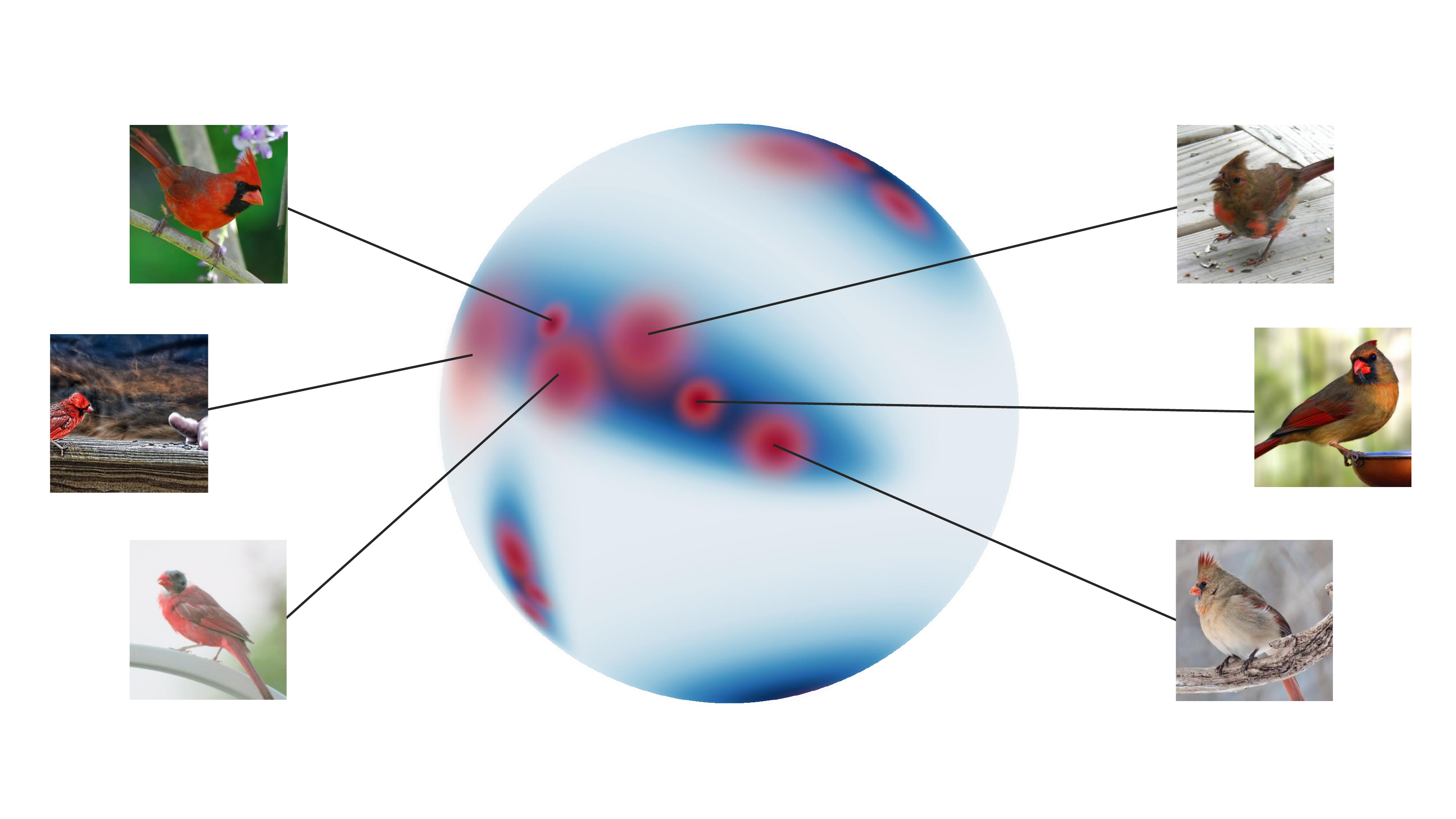}
    \caption{Class proxy distributions (\blue{blue}) and image distributions (\red{red}) embedded on the 3D unit sphere. The central proxy has a non-isotropic variance, so it can represent the high variance in body color between male (left) and female (right) cardinals and the low variance in their beak shape (top to bottom). Ambiguous images (e.g. middle left) have higher variance than images that clearly show class-discriminating features (top left, middle right). Best viewed in color.}
    \label{fig:fig1}
\vspace{-8pt}
\end{figure}

\section{Introduction}
Understanding and encoding visual similarity is a key concept that drives applications ranging from image (video) retrieval \cite{margin,npairs,angular,Kim_2021_CVPR,Brattoli_2020_CVPR} to clustering \cite{grouping} and face re-identification \cite{face_verfication_inthewild,semihard,sphereface,arcface}. Most commonly, approaches leverage Deep Metric Learning (DML) \cite{semihard,npairs,proxynca,margin,roth2020revisiting} to reformulate visual similarity learning into a surrogate, contrastive representation learning problem: Here, a deep network is tasked to embed images such that a simple predefined distance metric over pairs of embeddings represents their actual semantic relations.
Similar contrastive learning is used for representation learning tasks s.a. supervised image classification \cite{khosla2020supervised} or self-supervised learning \cite{moco,chen2020simple}.
Common DML approaches are formulated as ranking tasks over data tuples (e.g. pairs \cite{contrastive}, triplets \cite{semihard} or quadruplets \cite{quadtruplet}) of similar and dissimilar samples. Unfortunately, the complexity of sampling such tuples grows exponentially with the tuples size \cite{margin}. 
This has motivated recent advances in DML to focus on \textit{proxy-based} approaches, where the similar samples are summarized into learnable proxy representations \cite{proxynca,softriple} against which the sample embeddings are contrasted. 

While this allows for fast convergence and reliable generalization, drawbacks may arise both in the treatment of proxies and samples:
Firstly, the deterministic treatment of sample representations does not offer any degrees of freedom to address ambiguities and uncertainty (e.g., an image of a bird covered by branches). 
Secondly, isotropic distance scores between proxy-sample pairs (e.g., cosine similarity) provide only limited tools for the network to derive the similarity of samples within a class, as the distance to each proxy alone is insufficient to resolve relative sample placements around a proxy. 
This hinders class-specific variance and substructures to be successfully accounted for, which have been shown to notably benefit downstream generalization performance \cite{roth2020revisiting,milbich2020diva}. 

To address these issues, we propose a \textit{probabilistic} interpretation of proxy-based DML. Driven by the fact that modern DML consistently operates on hyperspherical (i.e., normalized) representations \cite{margin,roth2020revisiting}, we derive hyperspherical von-Mises Fisher (vMF) distributions for each sample. A sample embedding's direction controls the placement on the hypersphere, and therefore its semantic content, and its norm parametrizes the certainty of the distribution. 
In conjunction, we also treat class proxies probabilistically, but through \textit{non-isotropic} vMF distributions. 
This enforces the distributional prior over each class proxy to explicitly account for different, non-isotropic distributions, capturing more complex class-specific sample distributions (c.f. Figure \ref{fig:fig1}).
As this moves the DML training from point-based to distributional comparisons, we merge both components into a sound setup by motivating distribution-to-distribution matching metrics based on probabilistic product kernels.
Our full framework is supported through an extensive set of derivations and experimental ablations that showcase and support how the extension to probabilistic proxy-based DML offers significant improvements, with competitive performance across the standard DML benchmarks -- CUB200-2011 \cite{cub200-2011}, CARS196 \cite{cars196}, and Stanford Online Products \cite{lifted} -- even when compared to much more complex training methods.

Overall, our contributions can be summarized as:
\textbf{(1)} We propose and derive a novel probabilistic interpretation of proxy-based DML to account for sample and class ambiguities by reformulating the standard proxy-based metric learning approach to a distributional one on the hypersphere.
\textbf{(2)} We extend the vMF model to a non-isotropical one for each class proxy to better incorporate and address intra-class substructures for better generalization.
\textbf{(3)} We introduce various distribution-to-distribution metrics for DML and contrast them to traditional point-to-point metrics.
\textbf{(4)} We support our proposed framework through various derivational and experimental ablations showcasing how a distributional treatment can positively impact the learned representation spaces.
\textbf{(5)} Finally, we benchmark against standard DML approaches and provide further significant experimental support for our probabilistic approach to proxy-based DML.

\section{Related Work}

\textbf{Deep Metric Learning} comprises several conceptually different approaches. 
Firstly, one can define ranking tasks over data tuples such as pairs \cite{contrastive,margin}, triplets \cite{semihard}, quadruplets \cite{quadtruplet} or higher-order variants \cite{lifted,npairs,multisimilarity}. An underlying network then learns to solve each tuple presented by learning a representation space in which distances between embeddings correctly reflect their respective semantics/labelling. However, as the sizes of presented tuples increase, so does the tuple space each ranking task is sampled from, resulting in notable redundancy and impacted convergence behaviour \cite{semihard,margin,roth2020revisiting}. 
As a result, a secondary branch evolved focusing on heuristics which target ranking tuples fulfilling a set of predefined \cite{semihard,margin,multisimilarity,easypositive} or learned \cite{smartmining,roth2020pads} criteria. In a similar vein, DML research has also tried to address the sampling complexity issue through the replacement of tuple components with learned concept representations denoted as \textit{proxies}, with some approaches leveraging proxies in a classification-style setting \cite{zhai2018classification,arcface} or in a ranking fashion, where each sample is contrasted against a respective proxy \cite{proxynca,proxyncapp,kim2020proxy,softriple}.
Finally, benefits have also been found in orthogonal extensions and fundamental improvements to the general DML training pipeline, through various different approaches such as the usage of adversarial training \cite{daml}, synthetic samples \cite{dvml,hardness-aware}, higher-order or curvilinear metric learning \cite{horde,chen2019curvilinear}, feature mining for ranking \cite{mic,sharing,milbich2020diva} or proxy-based \cite{Roth_2022_NIR} approaches, a breakdown of the overall metric space into subspaces \cite{Sanakoyeu_2019_CVPR,bier,abier}, orthogonal modalities \cite{Roth_2022_Lang} or knowledge distillation \cite{s2sd}. 
Our proposed probabilistic proxy-based DML falls into this line of work, but is orthogonal to these other approaches, as these extensions can be applied in a method-agnostic fashion.
In particular, we extend proxy-based DML by specifically accounting for sample and class ambiguity through a distributional treatment of samples and proxies, and by utilizing non-isotropic proxy distributions to encourage more complex intra-class distributions around each proxy, which has been shown to be beneficial for generalization \cite{roth2020revisiting,dml_ood_gen,Roth_2022_NIR}.\\
\textbf{Probabilistic Embeddings.} 
Various approaches to DML can already be framed from a more probabilistic standpoint, where softmax-based approaches on the basis of cosine similarities \cite{arcface,proxyncapp,zhai2018classification} can be seen as analytical class posteriors if each class assumes a von Mises-Fisher (vMF) distribution \cite{park2019discriminative,Zhe2018DirectionalSD}.
While these methods implicitly model classes as vMFs, probabilistic embedding approaches further model each sample as a distribution in the embedding space \cite{shi2019probabilistic,li2021spherical,scott2021mises}. 
This allows the model to express uncertainty when images are ambiguous. 
Recent works argue that this ambiguity is captured in the image embedding's norm \cite{scott2021mises,ranjan2017l2,li2021spherical}: \cite{scott2021mises} argues that the embedding of an image that shows many class-discriminative features of one class consists of several vectors that all point in the same direction, resulting in a higher norm. 
On this basis, \cite{scott2021mises,li2021spherical} pioneered the use of embedding direction and norm to model each image as a vMF distribution, in particular for supervised classification. 
Utilizing vMF distributions, we are the first to introduce a full probabilistic proxy-based DML framework, yielding distribution-to-distribution metrics. 
Additionally, we propose a non-isotropic vMF for proxy distributions, which allows us to represent richer class structures in the embedding space beneficial to generalization \cite{roth2020revisiting,milbich2020diva}. 



\section{Non-isotropic Probabilistic Proxy-based DML}

\subsection{A Probabilistic Interpretation of Proxy-based DML}
\label{sec:prelim}
In this section, we extend the common DML framework to a probabilistic one.
Fundamentally, DML aims to find embedding functions $e: \mathcal{X} \rightarrow \mathcal{E}$ from image $\mathcal{X}\subset\mathbb{R}^{H \times W \times 3}$ to $M$-dimensional metric embedding spaces $\mathcal{E}\subset \mathbb{R}^M$ such that a distance function $d: \mathcal{E} \times \mathcal{E} \rightarrow \mathbb{R}$ between embeddings $z_1 = e(x_1)$ and $z_2 = e(x_2)$ of images $x_1, x_2\in\mathcal{X}$ reflects the semantic relation between them. 
The embedding space $\mathcal{E}$ is chosen to be the $M$-dimensional unit hypersphere $\mathcal{E} = \mathcal{S}^{M-1}$, i.e. $\lVert z \rVert = 1$.
While an euclidean $\mathcal{E}$ might appear more natural, recent works in DML \cite{margin,multisimilarity,roth2020revisiting,s2sd,kim2020proxy} and other contrastive learning domains like self-supervised learning \cite{wang2020understanding,chen2020simple,moco,s-vae18} have seen significant benefits in a directional treatment through normalization of embeddings to the unit hypersphere. 
This can in parts be attributed to better scaling with increased embedding dimensions \cite{hypersphere} and semantic information being mostly directionally encoded \cite{ranjan2017l2}.
%
%
To learn the respective embedding space $\mathcal{E}$, DML commonly employs ranking objectives over sample tuples. 
Based on the class assignments for each sample, an embedding network is tasked to minimize distances between same-class samples while maximizing them when classes differ.
More recently, proxy-based approaches \cite{proxynca,proxyncapp,kim2020proxy,softriple} directly model the class assignments by introducing class representatives during training -- the proxies $\text{p}\in\mathcal{S}^{M-1}$. These are contrasted against the sample embeddings $e(x) = z$ using a NCA-like \cite{nca} formulation (ProxyNCA, \cite{proxynca}), which was slightly modified by \cite{proxyncapp} as a softmax-loss
\begin{equation}
\label{eq:pnca}
    \mathcal{L}_\text{NCA++} =  \log\frac{\exp(-d(\text{p}^*, z) / t)}{\sum_{c=1}^C \exp(-d(\text{p}_c, z) / t)} \,.
\end{equation}
Here, $\text{p}^*$ denotes the ground-truth proxy associated with $z$, $t$ a temperature, and $d$ a distance metric, most commonly the negative cosine similarity $d = -s$ with $s(p_c, z) = (p_c z) / (\lVert p_c \rVert \lVert z \rVert)$.
This implies a problematic assumption: Since only angles between samples and proxies are leveraged, class-specific distribution variances around each proxy cannot be accounted for. Second, the deterministic underlying network $e$ induces a Dirac delta distribution over sample representations \cite{sinha2020uniform}. This treats all the input data the same regardless of the level of ambiguity, not accounting for sample-specific uncertainties.

Therefore, we suggest to represent samples and proxies as random variables $Z$ and $\text{P}$ with densities $\zeta$ and $\rho$ on $\mathcal{S}^{M-1}$, which allows both samples and proxies to carry uncertainty context to address sample ambiguity while encouraging to account for more complex class distributions.
This converts the above loss to
\begin{equation}
\label{eq:pnca_distr}
    \mathcal{L} = \log \frac{\exp(-d(\rho^*, \zeta) / t)}{\sum_{c=1}^C \exp(-d(\rho_c, \zeta) / t)} .
\end{equation}
Below in \S\ref{sec:dml}, we discuss how precisely $\rho$ and $\zeta$ are parametrized, and in \S\ref{sec:distrdist}, we find a $d(\cdot,\cdot)$ suitable for distribution-to-distribution matching .


\subsection{Probabilistic Sample and Proxy Representations}
\label{sec:dml}
\textbf{Sample Representations.} A common distribution on $\mathcal{S}^{M-1}$ is the von Mises-Fisher (vMF) distribution \cite{fisher1953dispersion,mardia2009directional,zimmerman2021inversion}. It parametrizes the sample distribution $\zeta$ by a direction vector $\mathbf{\mu}_z \in \mathcal{S}^{M-1}$ that points towards the mode of the distribution and a concentration parameter $\kappa_z \in \mathbb{R}_{\geq 0}$ that controls the spread around the mode, where a higher $\kappa_z$ yields a sharper distribution. The density $\zeta$ of a vMF-distributed sample $Z \sim \text{vMF}(\mathbf{\mu}_z, \kappa_z)$ at a point $\tilde{z} \in \mathcal{S}^{M-1}$ is
\begin{equation}
    \zeta(\tilde{z}) = C_M(\kappa_z) \exp\left( \kappa_z \, s(\tilde{z}, \mathbf{\mu}_z) \right) \label{form:vmf}.
\end{equation}
$C_M$ is the normalizing function which we approximate in high-dimensions (see Supp.~\ref{sec:approx_cp}). 
The advantage of the vMF is a duality to the un-normalized image embeddings $z=e(x) \in \mathbb{R}^M$: The natural parameter of the vMF is $\nu_z = \kappa_z \mathbf{\mu}_z \in \mathbb{R}^M$, such that if we set $\mu_z = \frac{z}{\lVert z \rVert}$ and $\kappa_z = \lVert z \rVert$, the embedding norm gives the vMF concentration without needing to explicitly predict it (as necessary for normal distribution \cite{chun2021probabilistic,shi2019probabilistic}).
This is further motivated by recent findings indicating that CNNs encode the amount of visible class discriminative features in the norm of the embedding (e.g. \cite{scott2021mises}). We validate this assumption in \S\ref{sec:norms}.
\begin{figure*}[t]
    \centering
    \captionsetup[subfigure]{oneside,margin={0.8cm,0cm}}
    \begin{subfigure}{0.42\textwidth}
    \centering
        \includegraphics[width=\linewidth, trim={2.5cm 0.7cm 3.3cm 1.8cm}, clip]{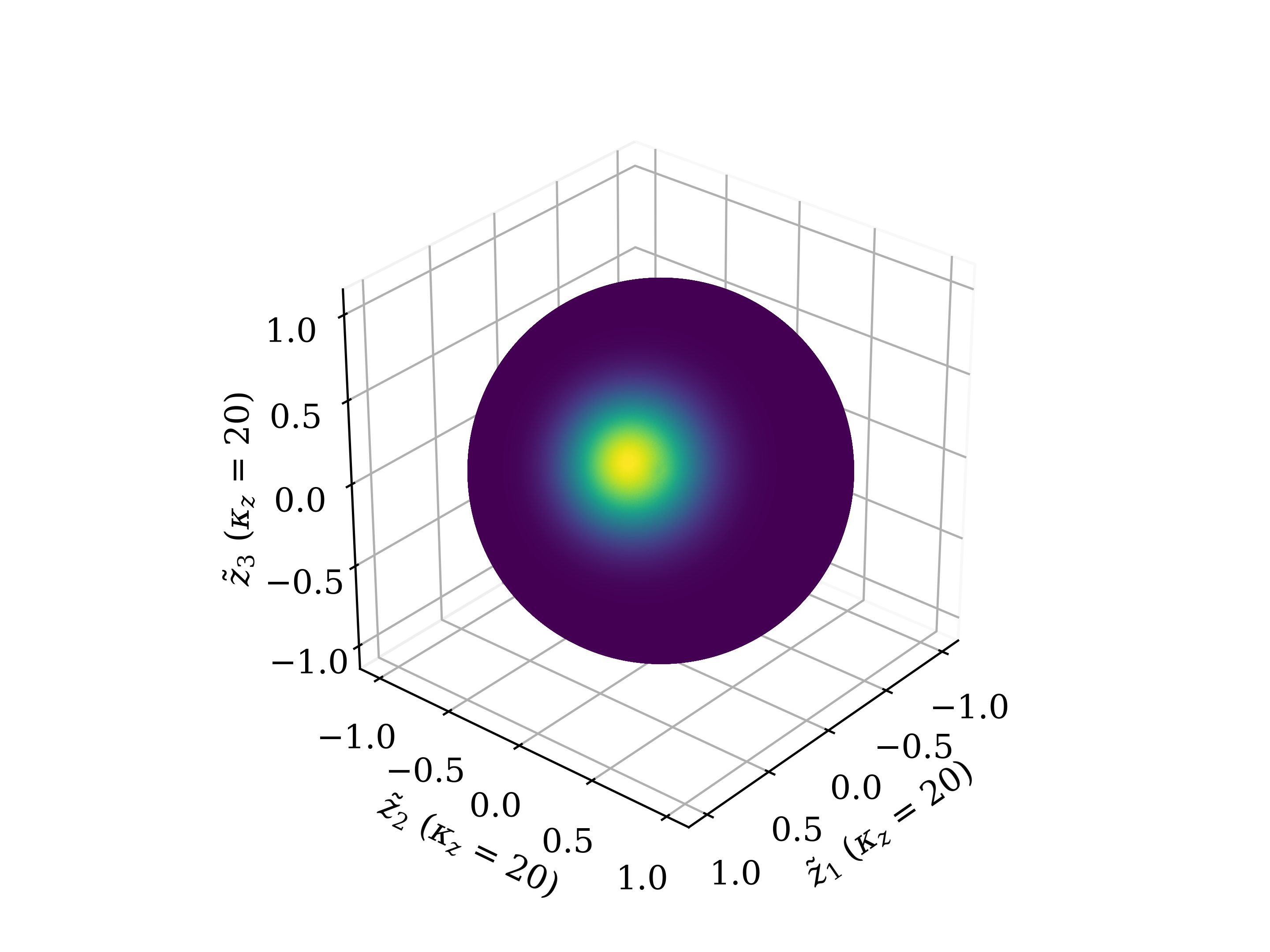}
        \caption{vMF, $\kappa_z = 20$} 
        \label{fig:nivmfa}
    \end{subfigure}%
    \begin{subfigure}{0.42\textwidth}
        \centering
        \includegraphics[width=\linewidth, trim={2.5cm 0.7cm 3.3cm 1.8cm}, clip]{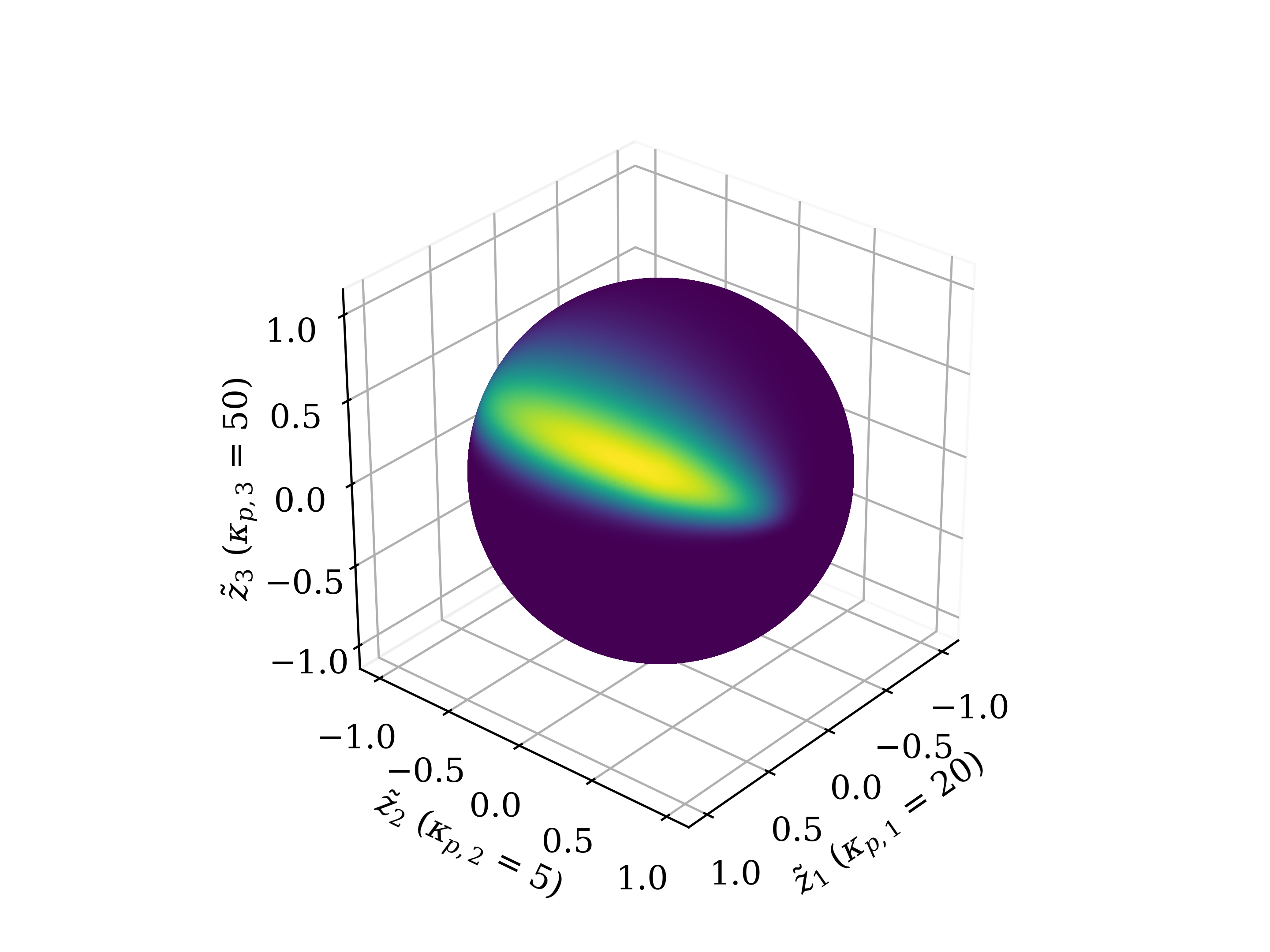}
        \caption{nivMF, $\mathbf{\kappa}_p = (20, 5, 50)$} 
        \label{fig:nivmfb}
    \end{subfigure}%
    \caption{Densities of (a) vMF and (b) non-isotropic vMF distributions on $\mathcal{S}^2$. The density is proportional to the color gradient from violet (zero) to yellow (high).}
    \label{fig:nivmf}
\vspace{-5pt}
\end{figure*}

\textbf{Proxy Representations.} It is possible to analogously treat the proxy distributions $\rho$ as vMF distributions with parameters $\nu_{\rho} = \kappa_{\rho} \mu_{\rho}$. However, a limiting factor owed to the simplicity of the vMF is its isotropy: The vMF is equivariant in all directions as shown in Figure~\ref{fig:nivmfa}. Proxies, however, need to account for more complex class distributions, i.e., non-isotropic ones (c.f. Fig.~\ref{fig:nivmfb}).
Generalized families of vMF distributions, such as Fisher-Bingham or Kent distributions \cite{mardia2009directional,mardia1975statistics,fisher_bingham}, are able to capture non-isotropy. However, they use covariance matrices with a quadratic number of parameters and constraints on their eigenvectors.
This complicates their training via gradient descent, especially in high dimensions. Hence, we propose a low-parameter vMF extension called non-isotropic von Mises-Fisher distribution (nivMF). %
Just like the vMF, the $M$-dimensional nivMF of a proxy $p$ is parametrized by a direction $\mathbf{\mu}_p \in \mathcal{S}^{M-1}$, but its concentration is described by a concentration \textit{matrix} $K_p \in \mathbb{R}^{(M \times M)}$. To reduce its parameters, we assume $K_p = \text{diag}(\mathbf{\kappa}_p) = \text{diag}(\kappa_{p, 1}, \dotsc, \kappa_{p, M})$ to be a diagonal matrix where $\kappa_{p, m} > 0, m = 1, \dotsc, M,$ gives the concentration per dimension. They are treated as learnable parameters (see Supp.\ref{sec:alg_summary}). Then, we define the density $\rho$ of a nivMF distributed proxy $\text{P} \sim \text{nivMF}(\mathbf{\mu}_p, K_p)$ at a point $\tilde{z} \in \mathcal{S}^{M-1}$ as
\begin{equation}
    \rho = f_P(\tilde{z}) := C_M(\lVert K_p \mathbf{\mu}_p \rVert) \, D(K_p) \, \exp\left( \lVert K_p \mathbf{\mu}_p \rVert \, s(K_p \tilde{z}, K_p \mathbf{\mu}_p) \right). \label{form:nivmf}
\end{equation}
with vMF normalizer $C_M$, and $D$ approximating an additional normalizing constant (see Supp.~\ref{sec:deriv_nivmf}). Intuitively, the nivMF is obtained from a vMF by a change-of-variable transformation: The unit sphere is stretched into an ellipsoid with axis lengths $\kappa_m, m=1, \dotsc, M,$ before the angle to the mode $\mathbf{\mu}_p$ is measured. Thus, distances of $\tilde{z}$ to $\mathbf{\mu}_p$ along dimensions with high concentrations are emphasized and distances along dimensions with low concentrations are weighted less. In effect, the $M$-dim $K_p$ is projected onto the $(M-1)$-dim tangential plane of $\mathbf{\mu}_p$ and controls the density's spherical shape (see Fig.~\ref{fig:nivmfb}). The remaining concentration projected on the $\mathbf{\mu}_p$-axis, i.e., $\lVert K_p \mathbf{\mu}_p \rVert$, controls the density's peakedness, analogously to the $\kappa$ parameter from a standard vMF. Thus, when $K_p = c I_M$ is the identity matrix scaled by some $c > 0$, the nivMF simplifies to a vMF (up to a constant due to an approximation, see Supp.~\ref{sec:deriv_nivmf}).

\subsection{Comparing Distributions Instead of Points}
\label{sec:distrdist}

As proxies and images are no longer modeled as points but as distributions, we present several distribution-to-distribution metrics (in the sense of distance functions $d$ in DML -- formally, they are no metrics as they don't fulfill the triangle inequality) and contrast them to traditional distribution-to-point metrics.

\textbf{Distribution-to-Distribution Metrics.} Probability product kernels (PPK) \cite{jebara2003bhattacharyya} are a family of metrics to compare two distributions $\rho$ and $\zeta$ by the product of their densities. One member of this family is the expected likelihood kernel (or mutual likelihood score \cite{shi2019probabilistic}). Although there is no analytical solution for nivMFs, we can derive a Monte-Carlo approximation
\begin{equation}
\label{form:elnivmf}
    d_{\text{EL-nivMF}}(\rho, \zeta) := -\log\left( \int_\mathcal{E} \rho(a) d\zeta(a) \right) 
    \approx -\log\left( \frac{1}{N} \sum\limits_{\substack{i=1, \dotsc, N \\ z_i \sim \zeta}} \rho(z_i) \right),
\end{equation}
where $N$ is the number of samples. Similar to \cite{scott2021mises}, we empirically found that a low number of samples ($N=5$) is sufficient. We use \cite{s-vae18} to sample from $\zeta$.

The expected likelihood kernel is advantageous since it is easily Monte-Carlo approximated, but there are other distribution-to-distribution metrics we would like to survey. Hence, we derive them under a vMF assumption for $\rho$, where they have analytical solutions (see Supp.~\ref{sec:more_metrics}). Namely, these are an analogous expected likelihood kernel $d_{\text{EL-vMF}}$, a related PPK kernel $d_{\text{B-vMF}}$, and a Kullback-Leibler distance $d_{\text{KL-vMF}}$. All three implicitly use the norm of the image embeddings in their calculations to respect the ambiguity, but differ in performance (see \S\ref{sec:ablation_metrics}). 


\textbf{Distribution-to-point Metrics.} Classical metrics like the cosine distance of the loss in Equation~\ref{eq:pnca} implicitly assume a distribution for each proxy and evaluate its log-likelihood at each sample. Hence, we will refer to them as distribution-to-point metrics. E.g., the cosine metric used in Equation~\ref{eq:pnca} is equivalent to the log-likelihood of the normalized sample embedding under vMF-distributed proxies with equal concentration values \cite{vmf_mix_1}, i.e., $d_{\text{Cos}}(\rho, \zeta) := -s(\mu_p, \mu_z) = -\log(\rho(\mu_z))$. Another common example is the L2-distance $d_{\text{L2}}(\rho, \zeta) := (\nu_p - \nu_z)^2 = -\text{log}(\rho(\nu_z))$ which is obtained by an equivariance normal distribution assumption for $\rho$. We analogously define $d_{\text{nivMF}}(\rho, \zeta) := -\text{log}(\rho(\mu_z))$ under a nivMF assumption for $\rho$ to benchmark it against the $d_{\text{EL-nivMF}}$ distance.

\subsection{Probabilistic Proxy-based Deep Metric Learning}
\label{subsec:probdml}
Utilizing distributional proxies $\rho$, distributional sample presentations $\zeta$ and the Monte-Carlo approximated Expected Likelihood Kernel $d_\text{EL-nivMF}(\rho, \zeta)$ we can fill in Equation~\ref{eq:pnca_distr} and define the probabilistic extension to proxy-based DML, precisely of the basic ProxyNCA (\cite{proxyncapp}), as
\begin{equation}
\label{eq:pnca_distr_2}
    \mathcal{L}^\text{EL-nivMF}_\text{NCA++} = \log \frac{\exp(-d_\text{EL-nivMF}(\rho^*, \zeta) / t)}{\sum_{c=1}^C \exp(-d_\text{EL-nivMF}(\rho_c, \zeta) / t)} \,.
\end{equation}
While this can be used as standalone loss, it can also probabilistically enhance other proxy-based objectives $\mathcal{L}_\text{Proxy-DML}$, such as ProxyAnchor \cite{kim2020proxy}. 
For easy usage in practice, we thus also propose using it as a regularizer via
\begin{equation}\label{eq:joint}
    \mathcal{L}_\text{joint}^\text{NCA++} = \mathcal{L}^\text{EL-nivMF}_\text{NCA++}(\rho, \zeta) + \omega\cdot\mathcal{L}_\text{Proxy-DML}(\mu_\rho, \mu_\zeta)
\end{equation}
with regularization scale $\omega$. Crucially, $\mu_\rho$ and $\mu_\zeta$ of the proxy and sample distributions are shared parameters with the non-probabilistic objective's proxies. This ensures alignment between the two learned representations spaces. 
The scaling $\omega$ balances the orthogonal benefits of the two approaches: An increasing $\omega$ highlights the non-probabilistic objective that encourages a better global alignment of distribution modes, and a decreasing $\omega$ yields a continuously more distributional treatment.
For the remainder of this work, we use \textbf{EL-nivMF} for the standalone probabilistic extension of ProxyNCA (Eq. \ref{eq:pnca_distr_2}), and $PANC$+\textbf{EL-nivMF} for the probabilistically regularized ProxyAnchor (Eq. \ref{eq:joint}).

\subsection{How Uncertainty-awareness Impacts Training}
\label{sec:norm_role}

\begin{figure}[t]
    \centering
    \begin{subfigure}{0.33\textwidth}
    \centering
        \includegraphics[width=\linewidth, trim={2.6cm 1.05cm 0.65cm 2.8cm}, clip]{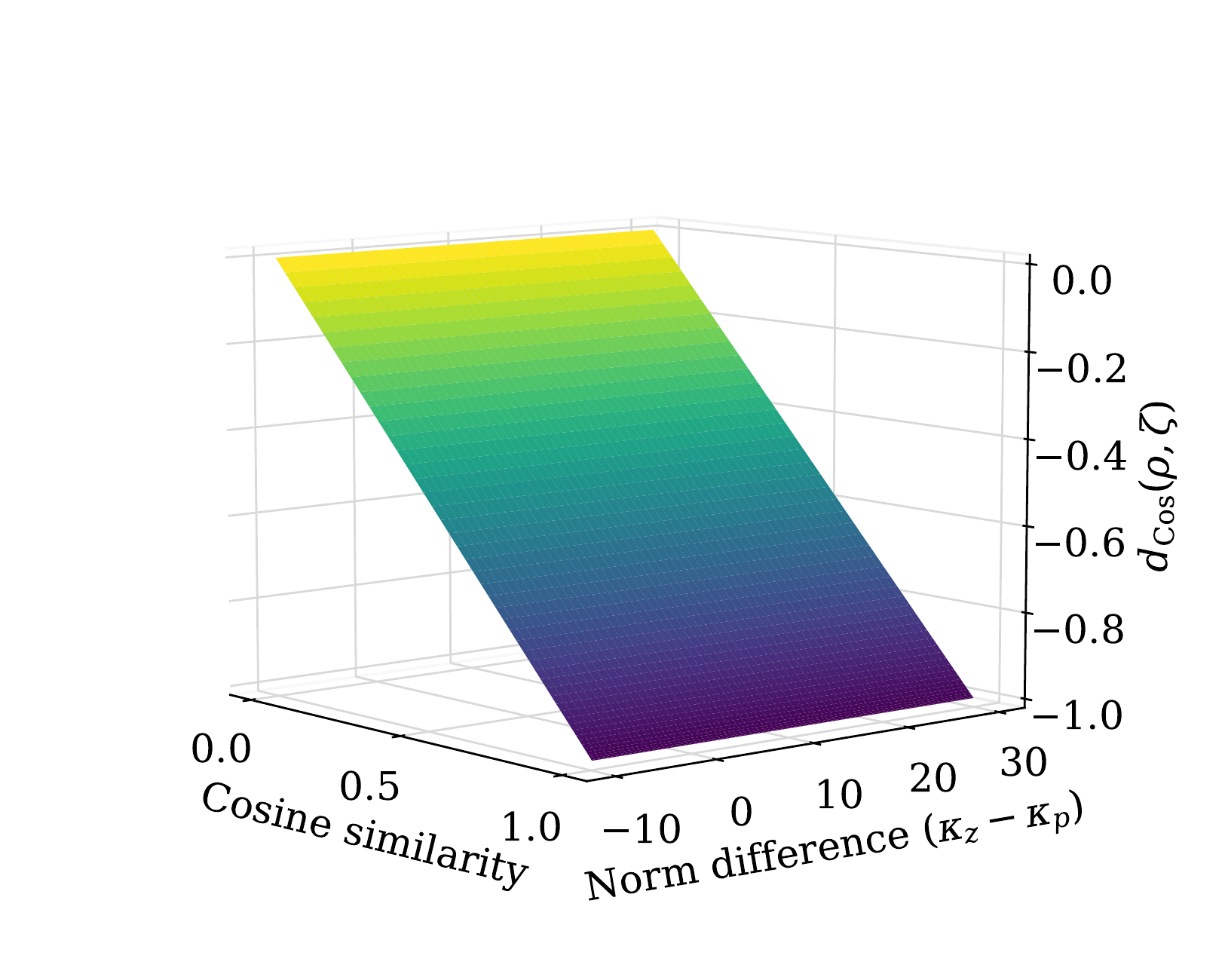}
        \caption{Cosine distance} 
        \label{fig:metric_cos}
    \end{subfigure}%
    \hfill{\hbox{}}   
    \begin{subfigure}{0.33\textwidth}
        \centering
        \includegraphics[width=\linewidth, trim={2.6cm 1.05cm 0.65cm 2.8cm}, clip]{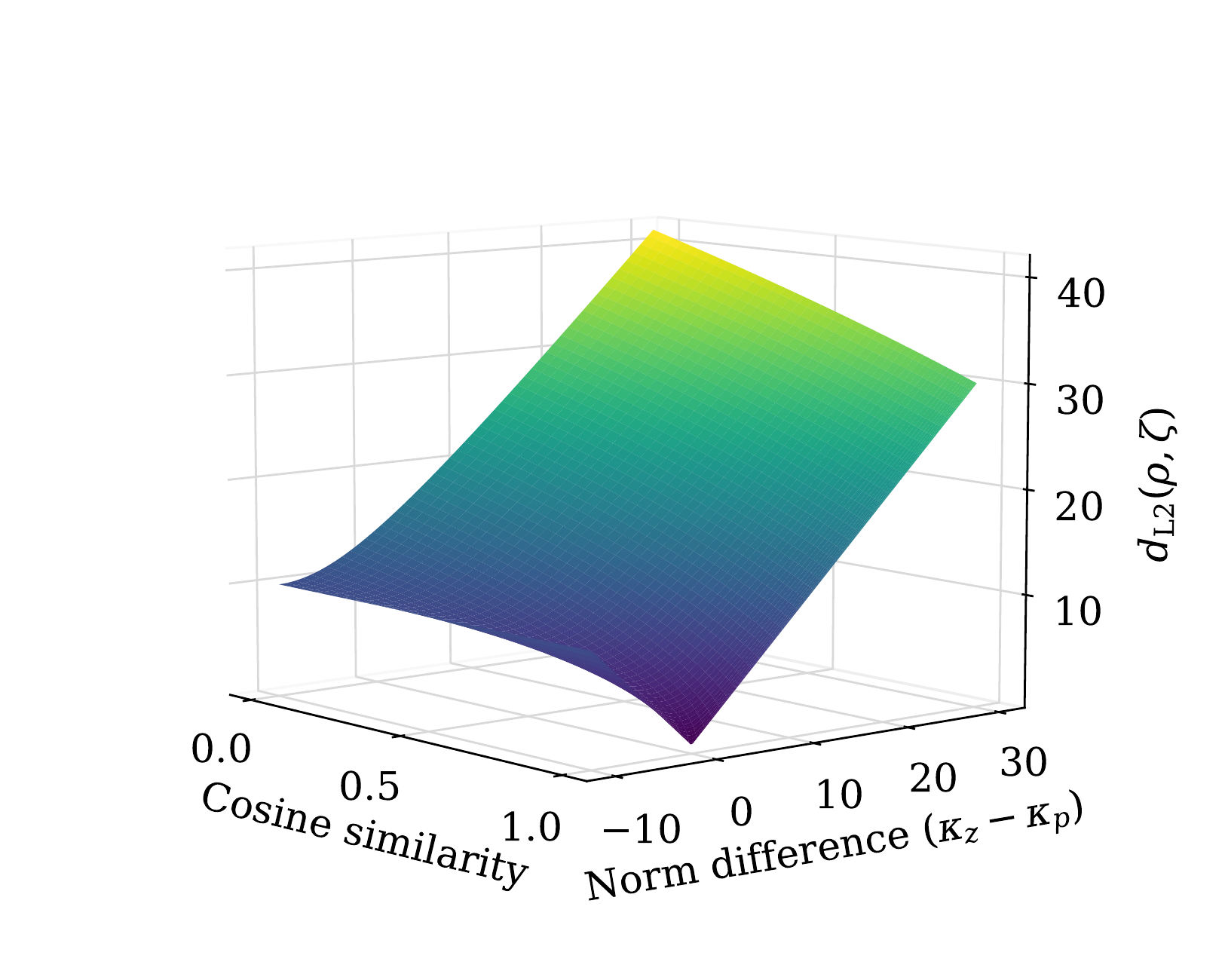}
        \caption{$L_2$ distance} 
        \label{fig:metric_l2}
    \end{subfigure}%
    \begin{subfigure}{0.33\textwidth}
    \hfill{\hbox{}}   
    \centering
        \includegraphics[width=\linewidth, trim={2.6cm 1.05cm 0.65cm 2.8cm}, clip]{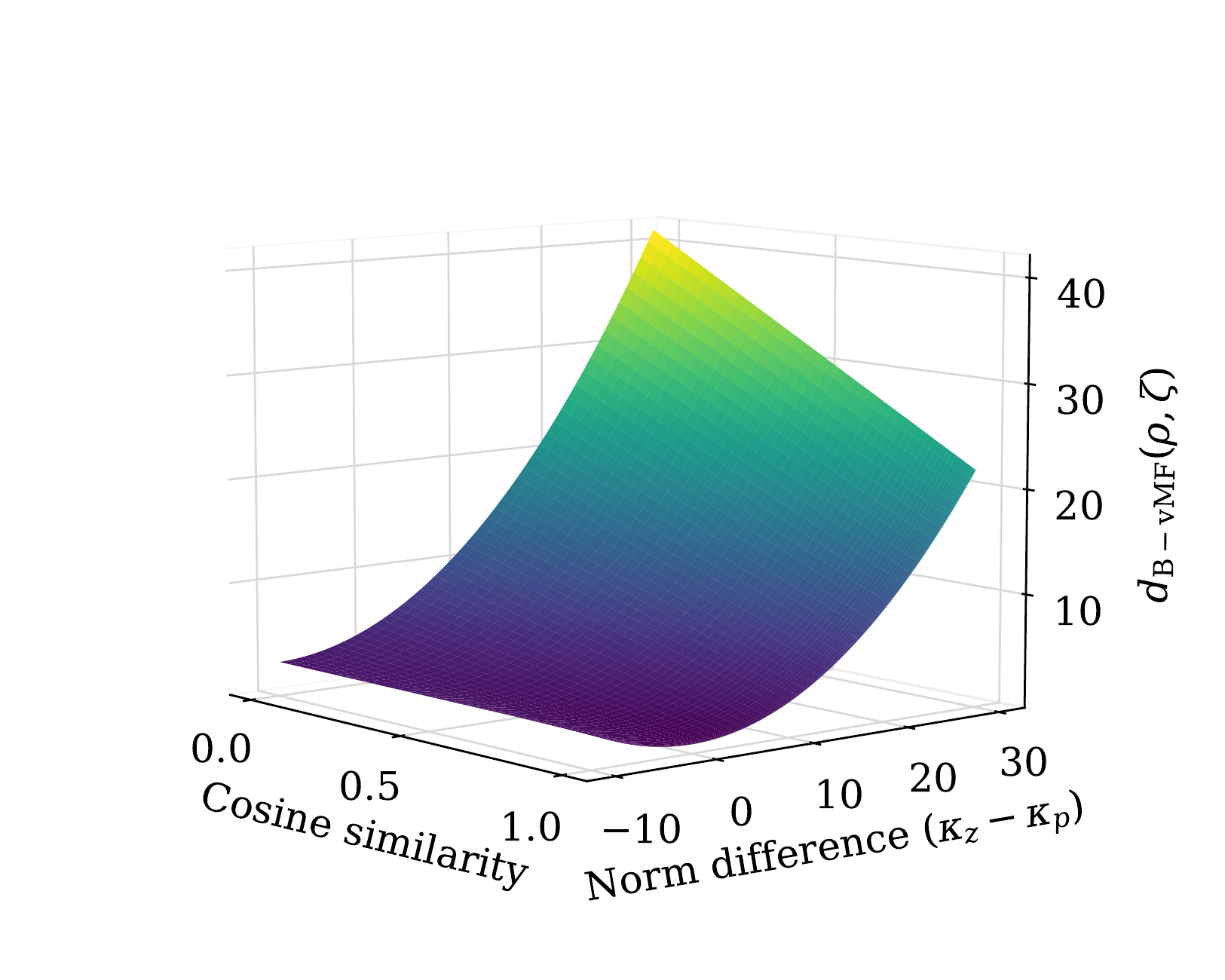}
        \caption{Bhattacharyya distance} 
        \label{fig:metric_bvmf}
    \end{subfigure}%
    \caption{Distances of a sample embedding to a vMF-distributed proxy with norm $\kappa_p=10$. (a) and (b) treat the sample as a point and (c) as a vMF distribution.}
    \label{fig:metric_comp}
\vspace{-6pt}
\end{figure}

Before the experimental evaluation, we provide an insight into \textit{how} incorporating uncertainty into the training benefits it. For this, we take a closer look at the norms of sample embeddings that, by duality, yield the concentration $\kappa_z$ of $\zeta$. 

\textbf{Uncertainty as Sample-wise Temperature.} Figure~\ref{fig:metric_comp} displays two distri-bution-to-point and one distribution-to-distribution metric with regard to the difference in norms and directions. We use the isotropic $d_{\text{B-vMF}}$ as a representative for distribution-to-distribution metrics since it has an analytical solution. While $d_{\text{Cos}}$ ignores the difference in norms, $d_{\text{L2}}$ and the similar, yet smoother, $d_{\text{B-vMF}}$ incorporate it
as an sample-wise temperature: The larger the norm of the sample gets, the steeper the metrics rise with increasing cosine distance. Thus, when comparing a sample to several proxies of roughly the same norm, their distances to the sample will be more uniform when the sample embedding norm is low and become more contrasted when it is high. In other words, ambiguous images produce more similar logits across all proxies and thus flatter class posterior distributions whereas highly certain images produce sharp posteriors.

\textbf{Uncertainty as Gradients Scale.} $\kappa_z$ has another influence on the training: Differentiating the losses $\mathcal{L}_{NCA++}^{\text{Cos}}$ and $\mathcal{L}_{\text{NCA++}}^{\text{L2}}$, obtained when using the norm-agnostic $d_{\text{Cos}}$ or the norm-aware $d_{\text{L2}}$ as distance functions in Eq.~\ref{eq:pnca}, w.r.t. the cosine similarity between $\mu_z$ and $\mu_p$ (as in \cite{kim2020proxy}) reveals (see Supp.~\ref{sec:grads})
\begin{align}
    \frac{\delta \mathcal{L}_{NCA++}^{\text{Cos}}}{\delta \cos(\mu_p, \mu_z)} &= \begin{cases}
    \frac{1}{t} \left(-1 + \frac{\exp(-d_{\text{Cos}}(\rho^*, \zeta)/t)}{\sum_{c=1}^C \exp(-d_{\text{Cos}}(\rho_c, \zeta)/t)} \right) & \text{if $p=p^*$} \\
    \frac{1}{t} \frac{\exp(-d_{\text{Cos}}(\rho^*, \zeta)/t)}{\sum_{c=1}^C \exp(-d_{\text{Cos}}(\rho_c, \zeta)/t)} & \text{else}
    \end{cases}\\ 
    \frac{\delta \mathcal{L}_{\text{NCA++}}^{\text{L2}}}{\delta \cos(\mu_p, \mu_z)} &= \begin{cases}
    \frac{2 \kappa_p \kappa_z}{t} \left( -1 + \frac{\exp(-d_{\text{L2}}(\rho^*, \zeta)/t)}{\sum_{c=1}^C \exp(-d_{\text{L2}}(\rho_c, \zeta)/t)} \right) & \text{if $p=p^*$} \\
    \frac{2 \kappa_p \kappa_z}{t} \frac{\exp(-d_{\text{L2}}(\rho^*, \zeta)/t)}{\sum_{c=1}^C \exp(-d_{\text{L2}}(\rho_c, \zeta)/t)} & \text{else}
    \end{cases},
\end{align}

where $p^*$ denotes the ground-truth class.
Besides the sample-wise temperature in $d_{\text{L2}}$, the gradients differ in that 
the gradient of $\mathcal{L}_{\text{NCA++}}^{\text{L2}}$ scales proportionally to $\kappa_z$. This means that in batch-wise gradient descent, samples with a high embedding norm are pulled towards ground-truth proxies and pushed away from others stronger than samples with low norm. In other words, the impact of an image on the structuring process of the embedding space depends on its ambiguity. This holds similarly for the distribution-to-distribution metrics, but is harder to derive than for $d_{\text{L2}}$. This analysis unveils that using the Euclidean $d_{\text{L2}}$ distance is adequate during training albeit switching to the hyperspherical $d_{\text{Cos}}$ at retrieval-time, as it can be seen as a simple approximation to the uncertainty-aware training of hyperspherical distribution-to-distribution metrics.

\section{Experiments}
We now detail the experiments (\S\ref{subsec:exp_details}) that benchmark our method (\S\ref{subsec:sota_comp}), before surveying different distr.-to-distr. metrics (\S\ref{sec:ablation_metrics}) and the role of the norm (\S\ref{sec:norms}).

\subsection{Experimental Details}\label{subsec:exp_details}
\textbf{Implementations.} All experiments use \href{https://pytorch.org/}{PyTorch} \cite{pytorch}. We follow standard DML protocols by leveraging ImageNet-pretrained ResNet50 \cite{resnet} and Inception-V1 networks with Batch-Normalization \cite{inceptionv1} as encoders. Their weights are taken from \href{https://pytorch.org/vision/stable/index.html}{torchvision} \cite{torchvision} and \href{https://github.com/rwightman/pytorch-image-models/tree/master/timm}{timm} \cite{timm}. 
To further ensure standardized training, we built upon the code and standardized DML protocols proposed in \cite{roth2020revisiting}, using the Adam optimizer \cite{adam}, a learning rate of $10^{-5}$ and weight decay of $4\cdot 10^{-3}$. In the more open state-of-the-art comparison (Table~\ref{tab:sota}), we additionally use step-wise learning rate scheduling. To ensure comparability and access to fast similarity search methods, all test-time retrieval uses cosine distances. To sample from vMF-distributions, we make use of \cite{s-vae18} and \href{https://github.com/google-research/vmf_embeddings}{respective implementations}. 
Further details on our method and hyperparameters are provided in Supp. \ref{supp:exp_details}. All experiments were run on NVIDIA 2080Ti GPUs with 12GB VRAM.\\
\textbf{Datasets.} We benchmark on three standard datasets: CUB200-2011 \cite{cub200-2011} (has a 100/100 split of train and test bird classes with 11,788 images in total), CARS196 \cite{cars196} (contains a 98/98 split of car classes and 16,185 images), and Stanford Online Products (SOP) \cite{lifted} (covers 22,634 product categories and 120,053 images).



\begin{table*}[t]
    \caption{We re-run various strong benchmarks in the \textit{standardized comparison} setting of \cite{roth2020revisiting}. We find strong improvements both when enhancing simple ProxyNCA towards probabilistic DML (\textbf{EL-nivMF}) and when using our approach as a regularizer on top of more versatile approaches (\textit{PANC} + \textbf{EL-nivMF}).}
 \footnotesize
  \setlength\tabcolsep{1.4pt}
  \centering
  \resizebox{1\textwidth}{!}{
  \begin{tabular}{l || c | c || c | c || c | c}
     \toprule
     \multicolumn{1}{l}{\textsc{Benchmarks}$\rightarrow$} & \multicolumn{2}{c}{\textsc{CUB200-2011}} & \multicolumn{2}{c}{\textsc{CARS196}} & \multicolumn{2}{c}{\textsc{SOP}} \\
     \midrule
     \textsc{Approaches} $\downarrow$ & R@1 & mAP@1000 & R@1 & mAP@1000 & R@1 & mAP@1000\\
    \midrule
    \hline
    \multicolumn{7}{l}{\textbf{Sample-based Baselines.}}\\
    \hline    
    Margin \cite{margin} & 62.9 $\pm$ 0.4 & 32.7 $\pm$ 0.3 & 80.1 $\pm$ 0.2 & 32.7 $\pm$ 0.4 & 78.4 $\pm$ 0.1 & 46.8 $\pm$ 0.1 \\    
    Multisimilarity \cite{multisimilarity} & 62.8 $\pm$ 0.2 & 31.1 $\pm$ 0.3 & 81.6 $\pm$ 0.3 & 31.7 $\pm$ 0.1 & 76.0 $\pm$ 0.1 & 43.3 $\pm$ 0.1 \\        
    \midrule
    \rowcolor{vlightgray}     
    \multicolumn{7}{l}{\textbf{Standard versus Probabilistic.}}\\
    ProxyNCA \cite{proxynca,proxyncapp} & 63.2 $\pm$ 0.2 & 33.4 $\pm$ 0.1 & 78.8 $\pm$ 0.2 & 31.9 $\pm$ 0.2 & 76.2 $\pm$ 0.1 & 43.0 $\pm$ 0.1 \\
    \textbf{EL-nivMF} & 64.8 $\pm$ 0.4 & 34.3 $\pm$ 0.3 & 82.1 $\pm$ 0.3 & 33.4 $\pm$ 0.2 & 76.6 $\pm$ 0.2 & 43.3 $\pm$ 0.1 \\    
    \midrule
    \rowcolor{vlightgray} 
    \multicolumn{7}{l}{\textbf{Probabilistic DML as Regularization.}}\\
    ProxyAnchor (\textit{PANC}, \cite{kim2020proxy}) & 64.4 $\pm$ 0.3 & 33.2 $\pm$ 0.3 & 82.4 $\pm$ 0.4 & 34.2 $\pm$ 0.3 & 78.0 $\pm$ 0.1 & 45.5 $\pm$ 0.1\\   
    \textit{PANC} + \textbf{EL-nivMF} & 66.5 $\pm$ 0.3 & 35.3 $\pm$ 0.1 & 83.6 $\pm$ 0.2 & 35.1 $\pm$ 0.1 & 78.2 $\pm$ 0.1 & 45.6 $\pm$ 0.1 \\  
    \bottomrule 

    \end{tabular}}
 \label{tab:relative_results}
 \vspace{-5pt}
\end{table*}

\begin{figure*}[t]
    \centering
    \includegraphics[width=1\textwidth]{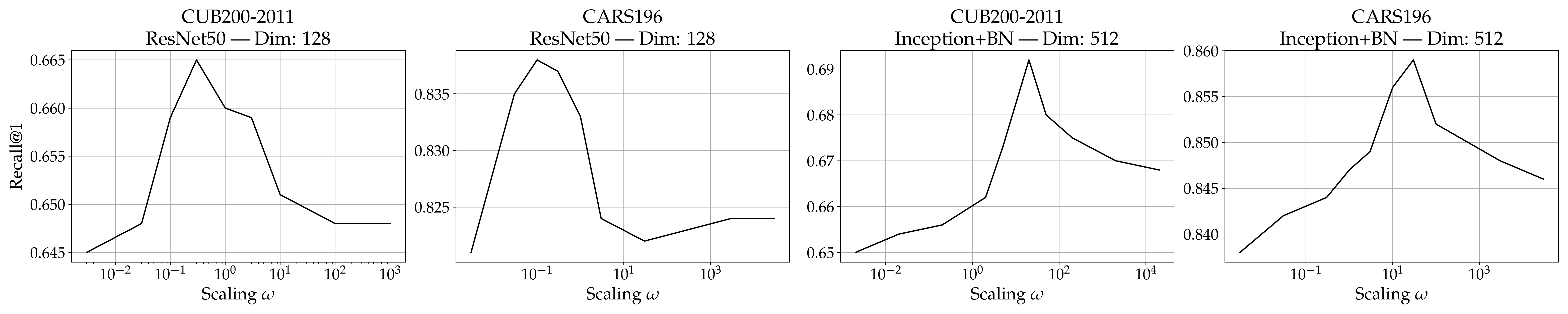}
    \caption{\textit{Probabilistic regularization as a function of the scaling factor} $\omega$. 
    We find a notable benefit when accounting for both orthogonal enhancements, i.e., the more probabilistic treatment (\textit{decreasing} $\omega$) and the better global alignment of the proxy distribution modes (\textit{increasing} $\omega$).}
\label{fig:scaling}
\vspace{-10pt}
\end{figure*}
\subsection{Quantitative evaluation of probabilistic proxy-based DML}\label{subsec:sota_comp}

\textbf{Standardized comparison.} We first follow protocols proposed in \cite{roth2020revisiting}, which suggest comparisons under equal pipeline and implementation settings (and no learning rate scheduling) to determine the true benefits of a proposed method, unbiased by external covariates. Particularily, we thus compare the standard ProxyNCA (see Eq. \ref{eq:pnca}) against our proposed \textbf{EL-nivMF} extension of ProxyNCA that includes sample and proxy distributions with distribution-to-distribution metrics during training. We further apply \textbf{EL-nivMF} as a probabilistic regularizer on top of the strong, but hyperparameter-heavy ProxyAnchor objective. Here, we only optimize the scaling $\omega$. Finally, we rerun the two strongest sample-based methods used in \cite{roth2020revisiting}.
In all cases, Table~\ref{tab:relative_results} shows significant improvements in performance and outperforms the sample-based methods. 
First, converting from standard to probabilistic proxy-based DML (ProxyNCA $\rightarrow$ \textbf{EL-nivMF}) increases R@1 on CUB200-2011 by $1.6pp$, $3.3pp$ on Cars196 and $0.4pp$ on SOP. 
This highlights the benefits of accounting for uncertainty and explicitly encouraging non-isotropic intra-class variance.
However, due to the large number of proxies and low number of samples per class, on SOP benefits are limited when compared to datasets such as CUB200-2011 and CARS196, as the estimation of our proxy distributions becomes noticeably noisier. 
When using \textbf{EL-nivMF} as probabilistic regularization, we find boosts of over $2.1pp$ and $1.2pp$ on CUB200-2011 and CARS196, respectively, with expected smaller improvements of $0.2pp$ on SOP.
Generally however, the consistent improvements, whether as a standalone objective or as a regularization method, highlight the versatility of a probabilistic take on DML, and offer a strong proof-of-concept for future DML research to built upon.

\textbf{Impact of different scaling factors $\omega$.}
Figure \ref{fig:scaling} showcases the generalization performance as a function of the scaling weight $\omega$ (see Eq. \ref{eq:joint}). Higher $\omega$ denotes a more non-probabilistic treatment to the point of ignoring the distributional aspects and returning to the auxiliary ProxyAnchor loss \cite{kim2020proxy}. Lower $\omega$ indicates a higher emphasis on distributional treatment of proxies (and samples).
Across benchmarks and backbones, the best performance is reached with an $\omega$ that is neither high nor $0$. Thus, the results highlight that our probabilistic proxy-based DML helps the better global realignment of each proxy distribution mode via ProxyAnchor, and vice-versa. Overall, R@1 increases up to $4pp$ at the most suitable scaling choice. This optimum is reached robustly in a large area around the peak (note the logarithmic x-axes).

\begin{table*}[t]
    \caption{\textit{Comparison to Literature}, separated by backbones and embedding dimensions. \textbf{Bold} denotes best results for a respective Backbone/Dim. subset, \blue{\textbf{bold}} the overall best. Results show that our probabilistically regularized ProxyAnchor method matches or beats previous, in parts notably more complex state-of-the-art methods.}
\setlength\tabcolsep{1.5pt}
\footnotesize
\centering

\resizebox{1\textwidth}{!}{
\begin{tabular}{l | l | l || c | c | c || c | c | c || c | c | c}
     \toprule
     \multicolumn{3}{l}{\textsc{Benchmarks} $\rightarrow$} & \multicolumn{3}{c}{\textsc{CUB200} \cite{cub200-2011}} & \multicolumn{3}{c}{\textsc{CARS196} \cite{cars196}} & \multicolumn{3}{c}{\textsc{SOP} \cite{lifted}}\\
     \midrule
     \textsc{Methods} $\downarrow$ & Venue & Arch/Dim. & R@1 & R@2 & NMI & R@1 & R@2 & NMI & R@1 & R@10 & NMI\\
     \midrule
     \hline
     Margin \cite{margin} & \textit{ICCV '17} &R50/128 & 63.6 & 74.4 & 69.0 & 79.6 & 86.5 & 69.1 & 72.7 & 86.2 & \textbf{90.7} \\
     Div\&Conq \cite{Sanakoyeu_2019_CVPR} & \textit{CVPR '19} & R50/128 & 65.9 & 76.6 & 69.6 & \textbf{84.6} & \textbf{90.7} & \textbf{70.3} & 75.9 & 88.4 & 90.2\\
     MIC \cite{mic} & \textit{ICCV '19} & R50/128 & 66.1 & 76.8 & 69.7 & 82.6 & 89.1 & 68.4 & 77.2 & 89.4 & 90.0\\
     PADS \cite{roth2020pads} & \textit{CVPR '20} &R50/128 & \textbf{67.3} & \textbf{78.0} & 69.9 & 83.5 & 89.7 & 68.8 & 76.5 & 89.0 & 89.9\\
     RankMI \cite{rankmi} & \textit{CVPR '20} & R50/128 & 66.7 & 77.2 & \textbf{71.3} & 83.3 & 89.8 & 69.4 & 74.3 & 87.9 & 90.5 \\
     \hline
    \rowcolor{vvlightgray} 
    \textit{PANC} + \textbf{EL-niVMF} & - & R50/128 & 67.0 & 77.6 & 70.0 & 84.0 & 90.0 & 69.5 & \textbf{78.6} & \textbf{90.5} & 90.1 \\             
    \hline
     NormSoft \cite{zhai2018classification} & \textit{BMVC '19} & R50/512 & 61.3 & 73.9 &  -   & 84.2 & 90.4 &  -   & 78.2 & 90.6 &  -    \\    
     EPSHN \cite{epshn} & \textit{WACV '20} & R50/512 & 64.9 & 75.3 &  -   & 82.7 & 89.3 &  -  & 78.3 & 90.7 &  -\\
     Circle \cite{circle} & \textit{CVPR '20} & R50/512 & 66.7 & 77.2 & - & 83.4 & 89.7 & - & 78.3 & 90.5 & - \\
     DiVA \cite{milbich2020diva} & \textit{ECCV '20} & R50/512 & 69.2 & 79.3 & 71.4 & \blue{\textbf{87.6}} & \blue{\textbf{92.9}} & 72.2 & 79.6 & \blue{\textbf{91.2}} & 90.6 \\
     DCML-MDW \cite{Zheng_2021_CVPR_compositional} & \textit{CVPR '21} & R50/512 & 68.4 & 77.9 & 71.8 & 85.2 & 91.8 & \blue{\textbf{73.9}} & \blue{\textbf{79.8}} & 90.8 & \blue{\textbf{90.8}} \\
     \hline
    \rowcolor{vvlightgray} 
    \textit{PANC} + \textbf{EL-niVMF} & - & R50/512 & \textbf{69.3} & \textbf{79.3} & \blue{\textbf{72.1}} & 86.2 & 91.9 & 70.3 & 79.4 & 90.7 & 90.6 \\       
    \hline
     Group \cite{elezi2020grouploss} & \textit{ECCV '20} &IBN/512 & 65.5 & 77.0 & 69.0 & 85.6 & 91.2 & \textbf{72.7} & 75.1 & 87.5 & \blue{\textbf{90.8}}   \\    
     DR-MS \cite{dutta2020orthogonalunsupdml} & \textit{TAI '20}& IBN/512 & 66.1 & 77.0 & - & 85.0 & 90.5 & -  & - & - & -   \\
     ProxyGML \cite{Zhu2020graphdml} & \textit{NeurIPS '20} & IBN/512 & 66.6 & 77.6 & 69.8 & 85.5 & 91.8 & 72.4  & 78.0 & 90.6 & 90.2   \\     
     DRML \cite{drml} & \textit{ICCV '21} & IBN/512 & 68.7 & 78.6 & 69.3 & \textbf{86.9} & \textbf{92.1} & 72.1 & 71.5 & 85.2 & 88.1 \\
     \textit{PANC} + \textit{MemVir} \cite{memvir} & \textit{ICCV '21} & IBN/512 & 69.0 & 79.2 & - & 86.7 & 92.0 & - & \textbf{79.7} & \textbf{91.0} & - \\
     \hline
    \rowcolor{vvlightgray} 
    \textit{PANC} + \textbf{EL-niVMF} & - & IBN/512 & \blue{\textbf{69.5}} & \blue{\textbf{80.0}} & \textbf{71.0} & 86.4 & 92.0 & 71.3 & 79.2 & 90.4 & 90.2 \\      
    \hline
     \bottomrule
\end{tabular}}

\vspace{-6pt}
\label{tab:sota}
\end{table*}
\textbf{Comparison against SOTA.} After these strictly standardized comparisons, we now compare the combination of ProxyAnchor and \textbf{EL-nivMF}, which performed best in the previous study to the larger DML literature. The hyperparameters and pipeline components (e.g., learning rate, weight decay) differ between the approaches, and so the comparison should be taken with a grain of salt \cite{roth2020revisiting,musgrave2020metric,scott2021mises}, but we still separate by the backbones and embedding dimensionalities, which are identified as the largest factors of variation \cite{roth2020revisiting}.
Accounting for that, we find competitive performance on all benchmarks (c.f. Table \ref{tab:sota}), even when compared against other, much more complex state-of-the-art methods relying on multitask learning (DiVA \cite{milbich2020diva}, MIC \cite{mic}) or reinforcement learning (PADS \cite{roth2020pads}). This makes our probabilistic take on proxy-based DML a generally attractive approach to DML, with further potential improvements down the line by implementing the probabilistic perspective into these orthogonal extensions.

\textbf{Computational overhead.} We do note that training with \textbf{EL-nivMF} requires the differentiable drawing of samples from vMF-distributions (see Eq.~\ref{form:elnivmf} and \cite{s-vae18}). This can increase the overall training time, but we found 2-5 samples to already be suitable, limiting the impact on overall walltime to $<25\%$ against pure ProxyNCA. This is in line with other extensions of ProxyNCA (s.a. \cite{horde,mic,Sanakoyeu_2019_CVPR,roth2020pads,milbich2020diva}). The retrieval walltime remains unaffected as cosine-similarity is deployed.
As an alternative for rapid training, we provide further probabilistic distribution-to-distribution distances ($d_\text{EL-vMF}$, $d_\text{B-vMF}$, $d_\text{KL-vMF}$) along with analytical solutions (Supp.~\ref{sec:more_metrics}), so that no sampling is required and computational overhead is negligible. We study them in the next section.
\begin{figure*}[t]
    \centering
    \begin{subfigure}{0.43\textwidth}
    \centering
        \includegraphics[width=\linewidth]{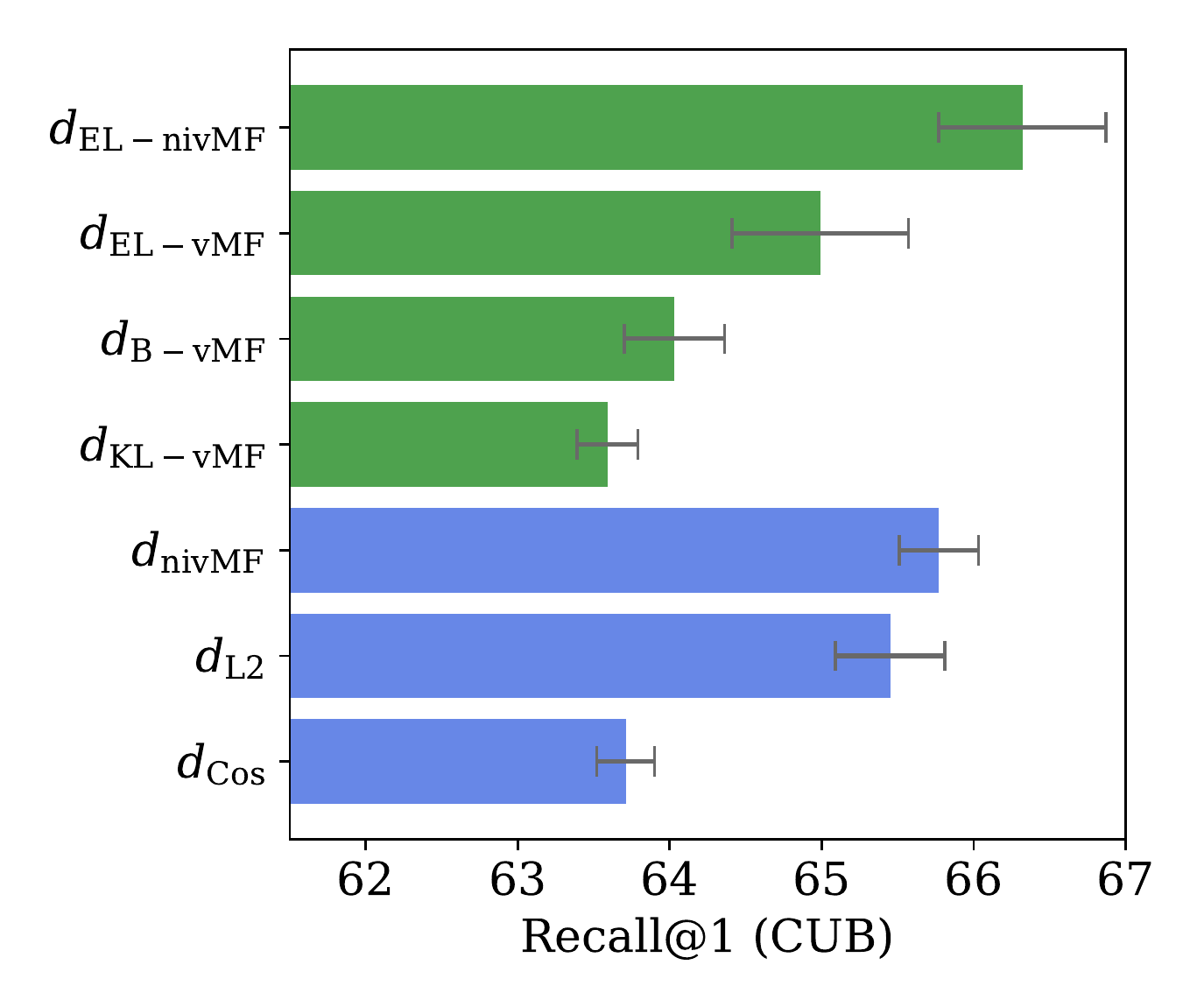}
    \end{subfigure}
    \begin{subfigure}{0.43\textwidth}
        \centering
        \includegraphics[width=\linewidth]{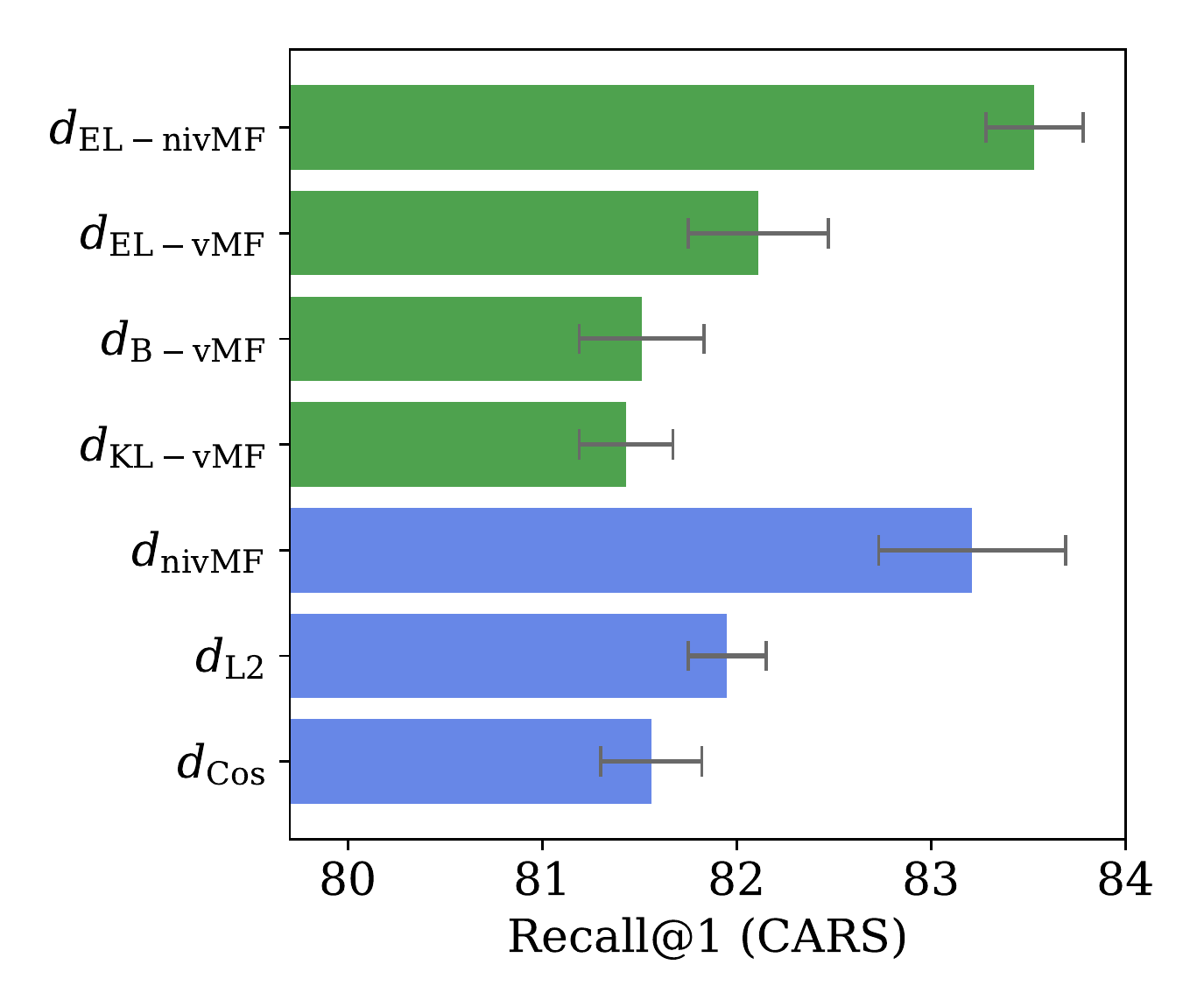}
    \end{subfigure}%
    \caption{Distance-to-point (blue) vs. distance-to-distance (green) metrics on CUB and CARS. Bars show average R@1 with standard deviation.}
    \label{fig:abl}
\vspace{-9pt}
\end{figure*}






\subsection{Quantitative Comparison of Metrics}\label{sec:ablation_metrics}
Sections \ref{sec:distrdist} and \ref{sec:dml} provided numerous modeling choices for distributions and distance metrics that can be plugged into the probabilistic DML framework in Eq.~\ref{eq:pnca_distr}. This section investigates these possibilities, ultimately motivating the particular choice of $d_\text{EL-nivMF}$, and also compares to more traditional distribution-to-point metrics. 
To ensure fair comparisons, we return to the standardized benchmark protocol of \cite{roth2020revisiting} using a 512-dimensional ResNet-50. All hyperparameters are fixed, except for the initial proxy norm and temperature, which are tuned via grid search on a validation set. 

Figure~\ref{fig:abl} shows the R@1 of all three distribution-to-point and four distribution-to-distribution metrics on CUB and CARS. Comparing the distribution-to-point metrics, $d_{\text{L2}}$ outperforms $d_{\text{Cos}}$ on both datasets, but is dominated by $d_{\text{nivMF}}$. The non-isotropic approach also performs best within the distribution-to-distribution metrics. Within the three isotropic distribution-to-distribution metrics, $d_{\text{KL-vMF}}$ shows the worst performance, with a small gap to the Bhattacharyya and a larger gap to the expected likelihood PPKs. This stands in line with preliminary findings of \cite{chun2021probabilistic}. The latter performs within one standard deviation of $d_{\text{L2}}$. 
Altogether, we find that adding non-isotropy to the standard $d_{\text{Cos}}$ (i.e., using $d_{\text{nivMF}}$) increases the R@1 by $2.1pp$ on CUB and $1.7pp$ on CARS. Further considering the image norm (i.e., $d_{\text{EL-nivMF}}$) adds another $0.6pp$ on CUB and $0.3pp$ on CARS.

The enhancement by non-isotropic modeling can be seen as inductive bias towards better resolution of intra-class variances and substructures (see Supp.~\ref{sec:furtherablations}), which drives generalization performance \cite{roth2020revisiting,milbich2020diva,dvml,drml,epshn}.
The strong performance of $d_{\text{L2}}$ is surprising as many current approaches use a $d_{\text{Cos}}$-based loss \cite{arcface,kim2020proxy,proxyncapp}. The crux is that $d_{\text{L2}}$ in our setting still uses the cosine distance at retrieval-time, similar to, e.g., \cite{boudiaf2020unifying}. 
Using $d_{\text{L2}}$ also as the retrieval metric would reduce the R@1 by up to $-5.34pp$ across all metrics and datasets, with the highest reduction appearing on the $d_{\text{L2}}$-trained model itself (see Supp.~\ref{sec:l2retrieval}). 
This supports the usage of the norm only during training, discussed in \S\ref{sec:norm_role}, where $d_{\text{L2}}$ shares the uncertainty-awareness of distribution-to-distribution metrics, explaining the small gap between $d_{\text{L2}}$ and $d_{\text{EL-vMF}}$. 
Thus, ultimately, we conjecture that it doesn't matter whether an approach is motivated from a distribution-to-distribution or distribution-to-point perspective, as long as it considers the ambiguity of images (and proxies) during training.

\subsection{Embedding Norms Encode Uncertainty}
\label{sec:norms}
\begin{figure*}[t]
    \centering
    \includegraphics[width=0.42\linewidth]{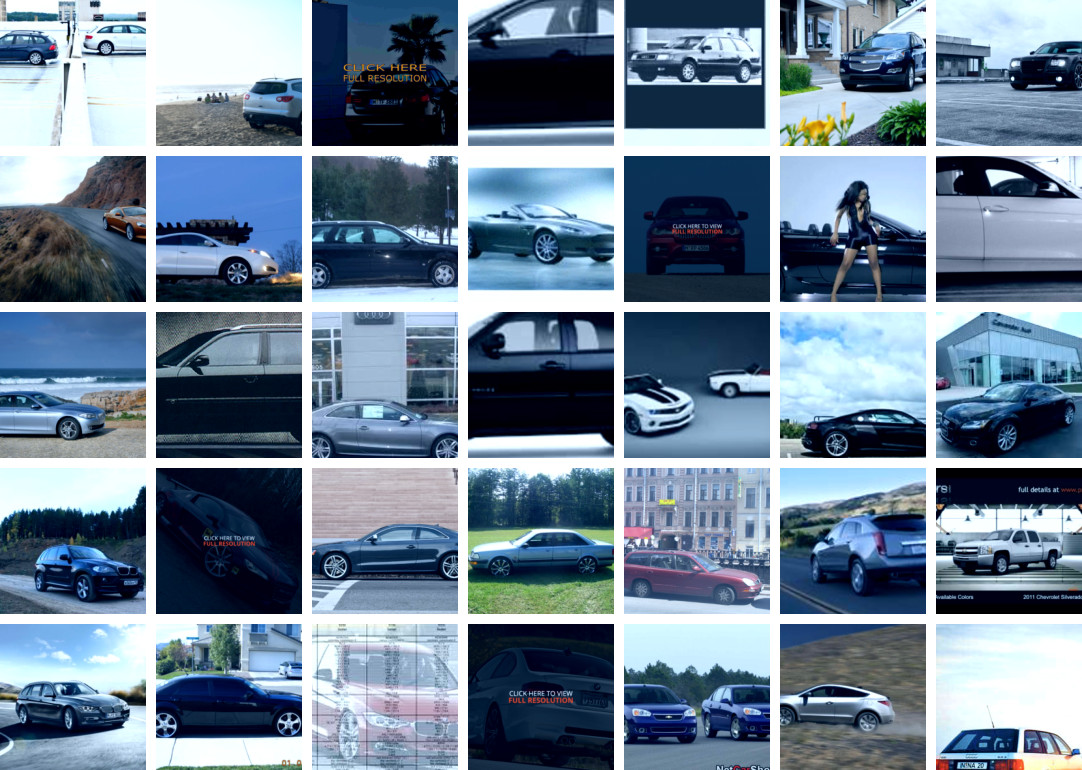}
    \begin{tikzpicture}
        \node (n1) {$\dots$}; 
        \node (n2) at ([yshift=0.8cm]n1) {$\dots$}; 
        \node (n3) at ([yshift=1.6cm]n1) {$\dots$}; 
        \node (n4) at ([yshift=2.4cm]n1) {$\dots$}; 
        \node (n5) at ([yshift=3.2cm]n1) {$\dots$}; 
    \end{tikzpicture}
    \includegraphics[width=0.42\linewidth]{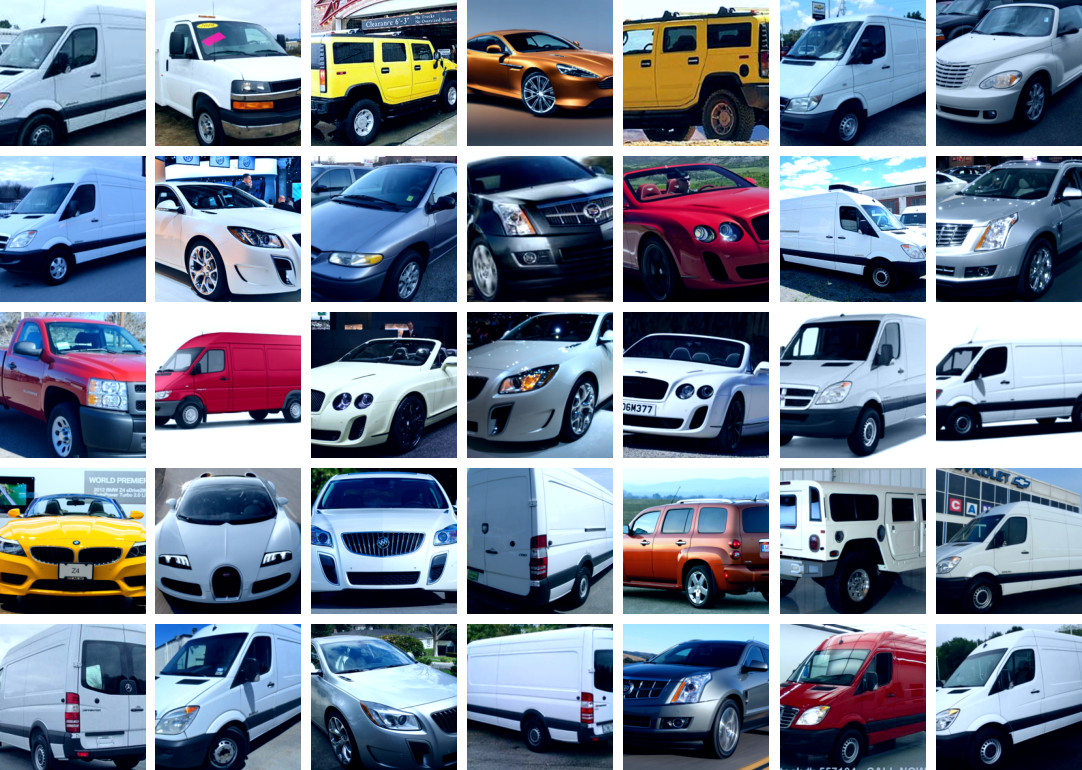}
    
    \begin{tikzpicture}
        \node (n1) {};
        \node (n3) [below of= n1, xshift=1cm, yshift=0.7cm]{lowest norm};
        \node (n2) at ([xshift=11.38cm]n1) {};
        \node (n4) [below of= n2, xshift=-1cm, yshift=0.7cm]{highest norm};
        \draw [-stealth] (n1) -- (n2);
    \end{tikzpicture}
    \caption{CARS train images with lowest (left) to highest (right) embedding norms.}
    \label{fig:qual_norm_global_ibn}
\vspace{-10pt}
\end{figure*}
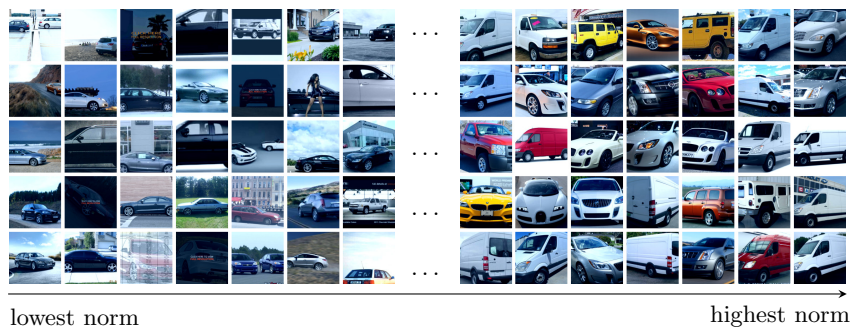

In the previous section, we found that considering the norms of embeddings during training leads to a higher performance. In this section, we qualitatively support that the learned norms actually correspond to a sample-wise ambiguity.

For this, we study the \textbf{EL-nivMF} model on CARS. 
Figure~\ref{fig:qual_norm_global_ibn} shows the images with the lowest and highest embedding norm in the training set. In many samples with low norm, characteristic parts of the cars are cropped out by the data augmentation (this also happens in the test set, hindering perfect accuracy). Others are overlaid or portray multiple distracting objects. In high-norm images, illumination and camera angle facilitate the detection of class-discriminative features. A competing hypothesis could be that high-norm images comprise mostly car classes with more distinctive designs. However, the differences between low and high norm images also hold within classes, see Supp.~\ref{sec:norm_hist} and \ref{sec:extra_norm}.
These findings are in line with \cite{scott2021mises,ranjan2017l2,li2021spherical} and support the hypothesis that the image norm indicates image certainties, motivated by being the sum of visible class-discriminative parts \cite{scott2021mises}. This justifies the $\kappa_z = \lVert z \rVert$ duality underlying the vMF assumption and is consistent with our analysis of uncertainty-aware training in \S\ref{sec:norm_role}.

%






\section{Conclusion}
This work proposes non-isotropic probabilistic proxy-based deep metric learning (DML) through uncertainty-aware training and non-isotropic proxy-distributions.
Uncertainty-aware training is achieved by treating sample embeddings not as deterministic points but as directional distributions parametrized by embedding directions and, beyond popular DML approaches, norms. 
This allows semantic ambiguities to be decoupled from the directional semantic context, which mathematically manifests itself in sample-wise temperature scaling and certainty-weighed gradients.
Additionally, our non-isotropic von Mises-Fisher distribution for proxies better models intra-class uncertainty, which introduces a low-parameter inductive prior for better generalizing embedding spaces. 
We support our approach through various ablation studies, which showcase that our proposed framework can operate both as a standalone objective and a probabilistic regularizer on top of existing proxy-based objectives. In both cases, we further found strong performances on the standard DML benchmarks, in parts matching or beating existing state-of-the-art methods. 
Our findings strongly indicate that a probabilistic treatment of proxy-based DML offers simple, orthogonal enhancements to existing DML methods and enables better generalization.\\
\textbf{Limitations.} We find that for applications with only few samples per class, the ability to estimate the non-isotropic proxy densities is limited (c.f. performance on SOP). 
For future work in such sparse settings, returning to the proposed isotropic distribution-to-distribution metrics or introducing across-class priors for the covariance matrices might serve as alternatives.

\subsection*{Acknowledgements}
This work has been partially funded by the ERC (853489 - DEXIM) and DFG (2064/1 – Project number 390727645) under Germany's Excellence Strategy.
Michael Kirchhof and Karsten Roth thank the International Max Planck Research School for Intelligent Systems (IMPRS-IS) for support.
Karsten Roth further acknowledges his membership in the European Laboratory for Learning and Intelligent Systems (ELLIS) PhD program.

\clearpage 

%
%
\bibliographystyle{splncs04}
\bibliography{main}

\renewcommand\thesection{\Alph{section}}

\def\ECCVSubNumber{3991}  

\title{Supplementary Material: \\ A Non-isotropic Probabilistic Take on Proxy-based Deep Metric Learning} 

\titlerunning{Non-isotropic Probabilistic Proxy-based DML}
%
\author{Michael Kirchhof\inst{1}$^{, *}$ \and
Karsten Roth\inst{1}$^{, *}$ \and
Zeynep Akata\inst{1} \and Enkelejda Kasneci\inst{1}}
\authorrunning{M. Kirchhof et al.}
%
\institute{$^1$University of Tübingen, Germany\\($^*$) equal contribution}

\setcounter{section}{0}
\setcounter{page}{20}
\setcounter{table}{3}
\setcounter{figure}{7}
\setcounter{equation}{10}

\section{Approximation of the von Mises-Fisher Distribution's Normalizing Constant}
\label{sec:approx_cp}

\begin{figure*}[hb]
    \centering
    \begin{subfigure}{0.24\textwidth}
        \centering
        \includegraphics[width=\linewidth]{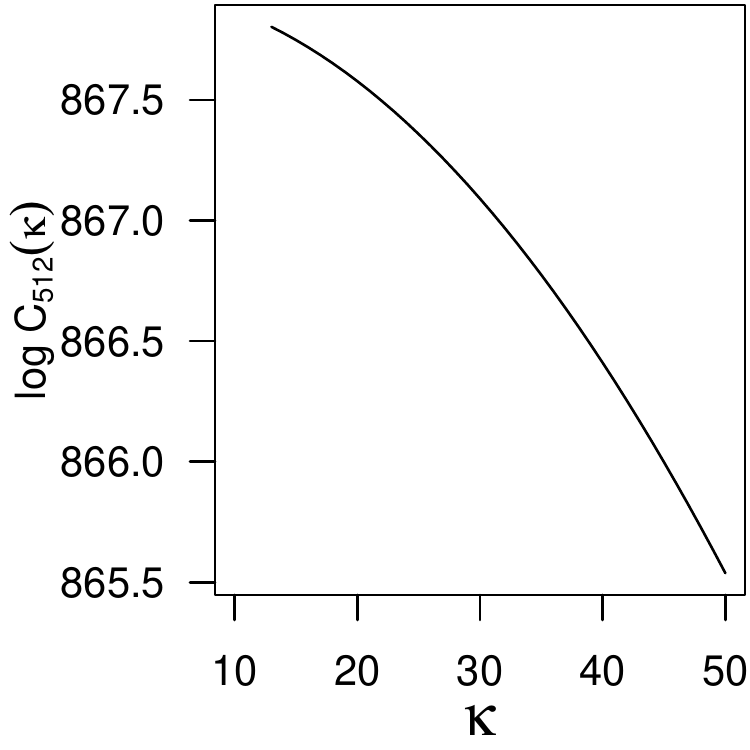}
        \caption{Exact values} 
        \label{fig:norm_consta}
    \end{subfigure}%
    \hfill{\hbox{}}   
    \begin{subfigure}{0.24\textwidth}
        \centering
        \includegraphics[width=\linewidth]{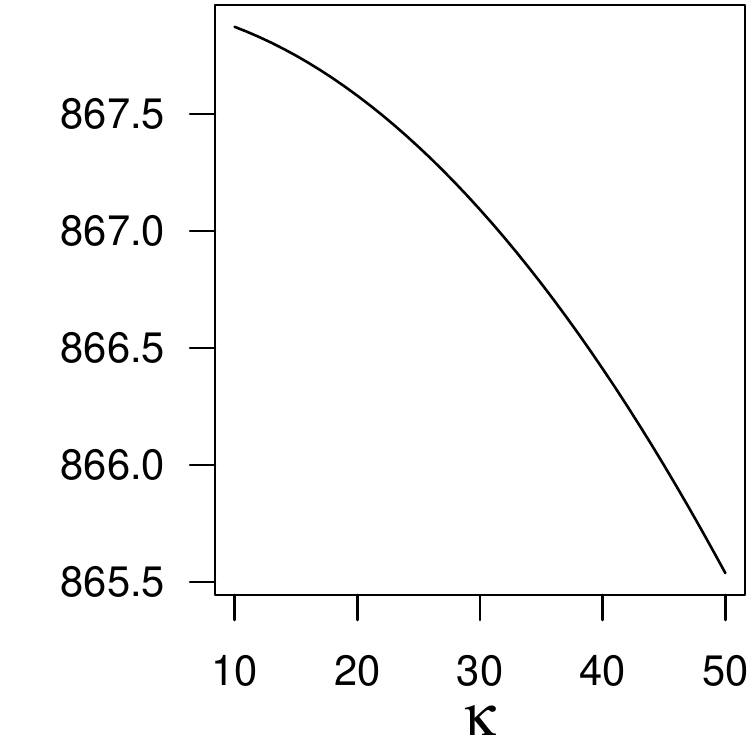}
        \caption{Our approx.} 
        \label{fig:norm_constb}
    \end{subfigure}%
    \hfill{\hbox{}}   
    \begin{subfigure}{0.24\textwidth}
        \centering
        \includegraphics[width=\linewidth]{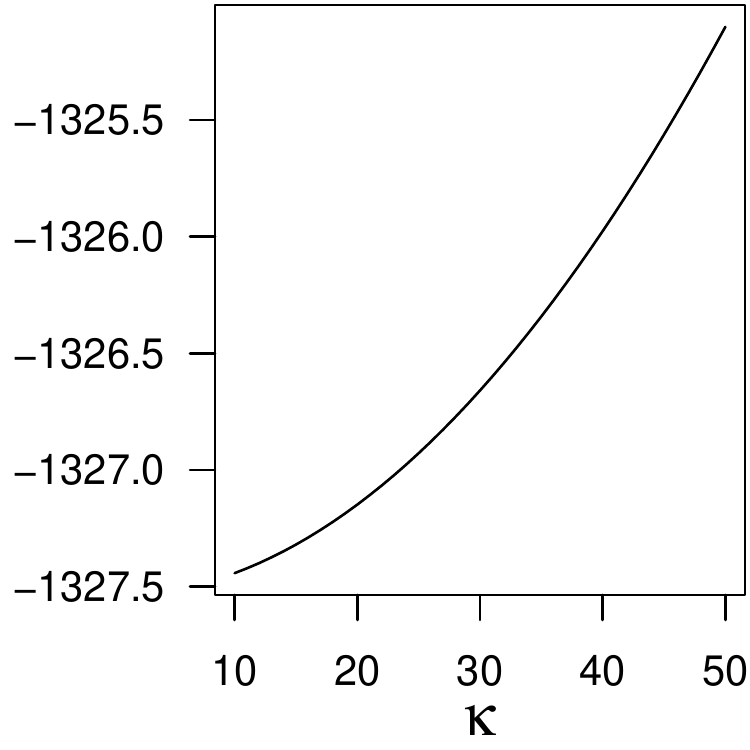}
        \caption{Approx. of \cite{kumar2018von}} 
        \label{fig:norm_constc}
    \end{subfigure}%
    \hfill{\hbox{}}   
    \begin{subfigure}{0.24\textwidth}
        \centering
        \includegraphics[width=\linewidth]{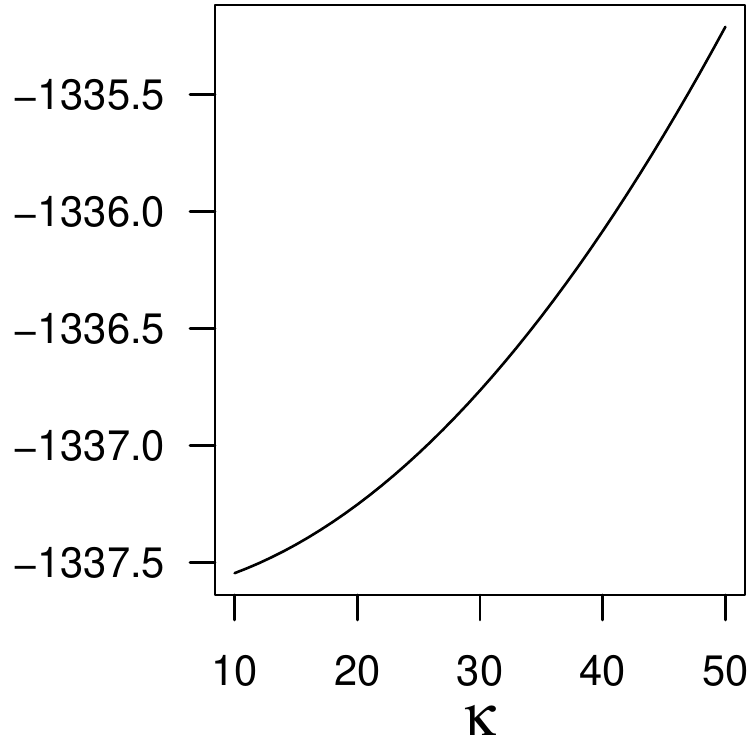}
        \caption{Approx. of \cite{scott2021mises}} 
        \label{fig:norm_constd}
    \end{subfigure}%
    \caption{Comparison of approximations and exact values of the logarithmized normalization constant of the vMF distribution $\log C_M(\kappa)$ for $M=512$ dimensions.}
    \label{fig:norm_const}
\end{figure*}

As we aim to resolve sample-specific ambiguities captured by $\kappa_z$, we need to calculate the logarithmic normalizing constant of the vMF distribution: 
\begin{equation}
    \log C_M(\kappa) = \log \frac{\kappa^{M/2 - 1}}{(2\pi)^{M/2} I_{M/2 -1}(\kappa)},
\end{equation}
where $I_d$ is the modified Bessel function of first kind at order $d$ and $M$ is the dimensionality of the embedding space. 
However, $I_d$ is expensive to compute and impossible to backpropagate through in high dimensions since it has no closed form. 
Hence, it is commonly approximated in the literature. \cite{kumar2018von} and \cite{scott2021mises} for example utilize approximations from lower and upper bounds which are shown in Figure~\ref{fig:norm_constc} and \ref{fig:norm_constd} for $M=512$. However, if we calculate $\log C_M$ from the exact Bessel functions implemented in \texttt{R 4.1.1}'s base package \cite{R411}, we see in Figure~\ref{fig:norm_consta} that $\log C_M$ is monotonically decreasing, because $I_d$ is monotonically increasing with $\kappa$ \cite[Section 10.37]{NIST}. 

To account for this issue, we thus choose to derive an approximation by directly fitting a quadratic model to the exact Bessel function for $M \in \{128, 512\}$ with $\kappa \in \{10, \dotsc, 50\}$. The resulting approximations are
\begin{align}
    \log C_{128}(\kappa) &\approx 127 - 0.01909 \cdot \kappa - 0.003355 \cdot \kappa^2 \text{ and} \\
    \log C_{512}(\kappa) &\approx 868 - 0.0002662 \cdot \kappa - 0.0009685 \cdot \kappa^2.
\end{align}
The mean squared error of these approximations to the ground truth values is smaller than $0.1\%$, which is visually confirmed in Figure~\ref{fig:norm_constb}. During experimentation, we found that the model is insensitive to perturbations in the precise coefficients. Also, we found that a linear model is too simple and an exponential model imposed very high gradients and inverts the behaviour of the metrics when $\kappa$ is high. Hence, we decided for the quadratic approximation as the simplest yet well extrapolating function. As a reference for future work, we note that \cite{kim2021pytorch} recently gave an additional approximation implemented in PyTorch.

\section{Derivation of the Non-isotropic von Mises-Fisher Distribution}
\label{sec:deriv_nivmf}
The nivMF can be motivated by a transformed vMF distribution, which we assume to be parametrized by $\mathbf{\mu} \in \mathcal{S}^{M-1}$ and $K = \text{diag}(\mathbf{\kappa}) \in \mathbb{R}^{(M \times M)}_{>0}, \mathbf{\kappa} \in \mathbb{R}^M_{>0}$. 
Transforming our parameters into $\mathbf{\tilde{\mu}} = \frac{K \mu}{\lVert K \mu \rVert}$ and $\tilde{\kappa} = \lVert K \mu \rVert$, we can define an ordinary vMF distribution $\tilde{X} \sim \text{vMF}(\mathbf{\tilde{\mu}}, \tilde{\kappa})$ with density
\begin{align}
    f_{\tilde{X}}(\tilde{x}) = C_M(\tilde{\kappa}) \, \exp \left( \tilde{\kappa} \tilde{x}^\top \mathbf{\tilde{\mu}} \right) \,\,.
\end{align}
For ease of notation, we do not include the subscript $p$ to denote specific proxies.
Now, we substitute $\tilde{x} := g(x) = \frac{K x}{\lVert K x \rVert}$. Note that $g$ is bijective as a function $g: \mathcal{S}^{M-1} \rightarrow \mathcal{S}^{M-1}$, but non-bijective when seen as a function $g: \mathbb{R}^M \rightarrow \mathbb{R}^M$, since it would lose a degree of freedom due to normalization. We will still treat it as the latter and ignore the non-bijectivity, such that the following should be seen as motivation and not proof, and comment on the implications further below. We now seek the density of $X = g^{-1}(\tilde{X})$. The change-of-variable theorem gives
\begin{align}
    f_X(x) &= f_{\tilde{X}}(\tilde{x}) \lvert \det \frac{\partial g(x)}{\partial x} \rvert \,. \label{form:changeofvariables}
\end{align}
By Equation 130 given in \cite{petersen2008matrix} and the chain rule, we obtain
\begin{align}
    \frac{\partial g(x)}{\partial x} &= \left( \frac{1}{\lVert K x \rVert} I_m - \frac{K^\top x x^\top K}{\lVert K x \rVert^3} \right) K^\top \\
    &= \left( \frac{1}{\tilde{\kappa}} I_M - \frac{(\tilde{\kappa} \mathbf{\tilde{\mu}}) (\tilde{\kappa} \mathbf{\tilde{\mu}})^\top}{\tilde{\kappa}^3} \right) K^\top \\
    &= \frac{1}{\tilde{\kappa}} \left( I_M - \mathbf{\tilde{\mu}} \mathbf{\tilde{\mu}}^\top \right) K^\top \,\,. \label{form:det}
\end{align}

Since the first part of this matrix is a projection on the orthogonal complement of $\mathbf{\tilde{\mu}}$, the matrix has rank $M-1$ and the determinant becomes zero. This is a consequence of the broken bijectivity assumption from above. However, we can see that Equation~\ref{form:det} essentially projects K on the tangential plane of $\mathbf{\tilde{\mu}}$. By taking its determinant, we measure the volume of the remaining $(M-1)$-dimensional concentration sphere. Performing a singular value decomposition on Equation~\ref{form:det} reveals that $\mathbf{\mu}$ is the eigenvector with eigenvalue 0. So, if we substract the contribution of $\mathbf{\mu}$ to the volume of $K$, which is $\lVert K \mathbf{\mu} \rVert = \tilde{\kappa}$, we obtain 
\begin{align}
    D(K) = \frac{\prod_{m=1}^M \kappa_m}{\tilde{\kappa}}\,.
\end{align} 
When we plug this heuristic into Equation~\ref{form:changeofvariables}, we arrive at the nivMF density:
\begin{align}
    f_X(x) &= C_M(\tilde{\kappa}) \, \exp \left( \tilde{\kappa} \tilde{x}^\top \mathbf{\tilde{\mu}} \right) D(K) \\
    &= C_M(\lVert K \mathbf{\mu} \rVert) \, D(K) \, \exp\left( \lVert K \mathbf{\mu} \rVert \, \left( \frac{K x}{\lVert K x \rVert} \right)^\top \frac{K \mathbf{\mu}}{\lVert K \mathbf{\mu} \rVert} \right) \\
    &= C_M(\lVert K \mathbf{\mu} \rVert) \, D(K) \, \exp\left( \lVert K \mathbf{\mu} \rVert \, s(K x, K \mathbf{\mu}) \right)\, .
\end{align}
We stress that $D(K)$ is a heuristic choice, such that the proposed nivMF density strictly speaking yields only a measure and not necessarily a probability measure. An analytical solution is promising material for future work. It may also enable the density of the nivMF to become a true expansion of the vMF density, i.e., $D(K)$ may vanish when $K = \kappa I_M$ for $\kappa > 0$, which is currently not the case. In empirical tests, dropping $D(K)$ lead to a considerably severed performance.

\section{Further distribution-to-distribution Metrics}
\label{sec:more_metrics}
We can define further distribution-to-distribution metrics beyond $d_{\text{EL-nivMF}}$. One starting point are probability product kernels (PPK) \cite{jebara2003bhattacharyya}. They are a family of metrics to compare two distributions $\rho$ and $\zeta$ by the product of their densities:
\begin{align}
    \text{PPK}_\gamma(\rho, \zeta) = \int_\mathcal{E} \rho(a)^\gamma \zeta(a)^\gamma da, \text{ with }\gamma > 0 .
\end{align}
Since the loss in Equation~\ref{eq:pnca_distr} takes the exponential of the distance metrics, we take their logarithms here to retain the PPK as actual score in nominator and denominator. In particular, if we assume a vMF distribution for both $\rho$ and $\zeta$
\begin{align}
    d_{\text{B-vMF}}(\rho, \zeta) := -\text{log}(\text{PPK}_{0.5}(\rho, \zeta))
\end{align}
gives the Bhattacharyya distance and
\begin{align}
    d_{\text{EL-vMF}}(\rho, \zeta) := -\text{log}(\text{PPK}_{1}(\rho, \zeta))
\end{align}
gives the expected likelihood distance, also known as mutual likelihood score \cite{shi2019probabilistic}. Their analytical solutions are provided in Supp.~\ref{sec:proofppk}. 

The previous metrics are symmetric in $\rho$ and $\zeta$. To capture the inherent asymmetry between samples and proxies, we also study the Kullback-Leibler divergence $d_{\text{KL-vMF}}(\rho, \zeta) := \text{KL}(\zeta||\rho)$. Its analytical solution if both $\rho$ and $\zeta$ are vMF densities is given in Supp.~\ref{sec:proofkl}. 

\section{Analytical Solutions of Bhattacharyya and Expected Likelihood Distance}
\label{sec:proofppk}

Let $\zeta$ and $\rho$ be densities of two vMF-distributed random variables with parameters $\nu_z = \kappa_z \mu_z$ and $\nu_p = \kappa_p \mu_p$, respectively.

\textbf{Bhattacharyya distance.} Since the vMF is a member of the exponential family, \cite{jebara2003bhattacharyya} gives us that 
\begin{align}
    \text{PPK}_{0.5}(\rho, \zeta) &= \exp(K(\nu_z/2 + \nu_p/2) - K(\nu_z)/2 - K(\nu_p)/2), \text{ with} \\
    K(\nu) &= - \log C_M(\lVert\nu\rVert) \,.
\end{align}
Thus, 
\begin{align}
    d_{\text{B-vMF}}(\rho, \zeta) &= -\text{log}(\text{PPK}_{0.5}(\rho, \zeta)) \\
    &= \log C_M(\lVert \nu_z + \nu_p \rVert / 2) - \log C_M(\nu_z)/2 - \log C_M(\nu_p)/2 \,.
\end{align}

\textbf{Expected likelihood distance.} We can extend
\begin{align}
    \text{PPK}_{1}(\rho, \zeta) &= \int_\mathcal{E} \zeta(\tilde{z}) \rho(\tilde{z}) d \tilde{z} \\
    &= C_M(\kappa_z) \cdot C_M(\kappa_p) \int_\mathcal{E} \exp((\kappa_z \mu_z + \kappa_p \mu_p)^\top \tilde{z}) d \tilde{z} \\
    &= \frac{C_M(\kappa_z) \cdot C_M(\kappa_p)}{C_M(\lVert \nu_0 \rVert)} \int_\mathcal{E} C_M(\lVert \nu_0 \rVert) \exp(\nu_0^\top \tilde{z}) d \tilde{z}, \text{ with} \\
    \nu_0 &:= \kappa_z \mu_z + \kappa_p \mu_p,
\end{align}
such that the latter is again the density of a vMF distributed random variable, whose integral over the embedding space is $1$. Then, 
\begin{align}
    d_{\text{EL-vMF}}(\rho, \zeta) &= -\text{log}(\text{PPK}_{1}(\rho, \zeta)) \\
    &= \log C_M(\lVert \nu_z + \nu_p \rVert) - \log C_M(\nu_z) - \log C_M(\nu_p) \,.
\end{align}
Note that both $d_{\text{EL-vMF}}$ and $d_{\text{B-vMF}}$ depend on $\lVert \nu_z + \nu_p \rVert$ which implicitely respects the cosine similarity between $\mu_z$ and $\mu_p$, but also processes $\kappa_z$ and $\kappa_p$.

\section{Analytical Solution of KL-Divergence}
\label{sec:proofkl}

Let $\zeta$ and $\rho$ be densities of two vMF-distributed random variables with parameters $\mu_z, \kappa_z$ and $\mu_p, \kappa_p$, respectively. Then
\begin{align}
    KL(\zeta||\rho) &= \int_\mathcal{E} \zeta(\tilde{z}) \log \frac{\zeta(\tilde{z})}{\rho(\tilde{z})} d \tilde{z} \\
    &= \int_\mathcal{E} \log C_M(\kappa_z) - \log C_M(\kappa_p) + (\kappa_z \mu_z^\top - \kappa_p \mu_p^\top) \tilde{z} d\zeta(\tilde{z}) \\
    &= \log C_M(\kappa_z) - \log C_M(\kappa_p)+ (\kappa_z \mu_z^\top - \kappa_p \mu_p^\top) \int_\mathcal{E} \tilde{z} d\zeta(\tilde{z}) \\
    &= \log C_M(\kappa_z) - \log C_M(\kappa_p)+ (\kappa_z \mu_z^\top - \kappa_p \mu_p^\top) \mu_z
\end{align}







\section{Gradients of $d_{\text{L2}}$ and $d_{\text{Cos}}$}
\label{sec:grads}


We are interested in differentiating the loss $\mathcal{L}_{\text{NCA++}}$ from Equation~\ref{eq:pnca} in \S\ref{sec:dml} by the cosine similarity between the image $z$ and a proxy of interest $p$. Let $p^*$ denote the ground-truth proxy of $z$ and $\frac{\delta}{\delta s} := \frac{\delta}{\delta s(\mu_p, \mu_z)}$. Then, 
\begin{align}
    \frac{\delta}{\delta s} \mathcal{L}_{\text{NCA++}} &= \begin{cases}
        \frac{\delta}{\delta s} d(\rho^*, \zeta)/t + \frac{\delta}{\delta s} \log(\sum_{c=1}^C \exp(-d(\rho_c, \zeta)/t)) &, \text{if $p=p^*$} \\
        \frac{\delta}{\delta s} \log(\sum_{c=1}^C \exp(-d(\rho_c, \zeta)/t)) &, \text{else} \label{form:diff_loss}
    \end{cases}
\end{align}
and by the chain rule we get
\begin{align}
    \frac{\delta}{\delta s} \log\left(\sum_{c=1}^C \exp(-d(\rho_c, \zeta)/t)\right) = -\frac{\exp(-d(\rho, \zeta)/t)}{\sum_{c=1}^C \exp(-d(\rho_c, \zeta)/t)} \frac{\delta}{\delta s} d(\rho_c, \zeta)/t\,\,. \label{form:diff_lse}
\end{align}

Let's consider the $\mathcal{L}_{\text{NCA++}}^{\text{Cos}}$ loss, i.e., $d(\rho, \zeta) = -s(\mu_p, \mu_z)$. We can plug $\frac{\delta}{\delta s} d(\rho, \zeta) = -1$ into Equations~\ref{form:diff_loss} and \ref{form:diff_lse} and obtain:
\begin{align}
    \frac{\delta}{\delta s} \mathcal{L}_{\text{NCA++}}^{\text{Cos}} &= \begin{cases}
    \frac{1}{t} \left(-1 + \frac{\exp(-d(\rho, \zeta)/t)}{\sum_{c=1}^C \exp(-d(\rho_c, \zeta)/t)} \right) &, \text{if $p=p^*$} \\
    \frac{1}{t} \frac{\exp(-d(\rho, \zeta)/t)}{\sum_{c=1}^C \exp(-d(\rho_c, \zeta)/t)} &, \text{else}
    \end{cases} \\
    &= \begin{cases}
    \frac{1}{t} \left(-1 + \frac{\exp(s(\mu_p, \mu_z)/t)}{\sum_{c=1}^C \exp(s(\mu_{p_c}, \mu_z)/t)} \right) &, \text{if $p=p^*$} \\
    \frac{1}{t} \frac{\exp(s(\mu_p, \mu_z)/t)}{\sum_{c=1}^C \exp(s(\mu_{p_c}, \mu_z)/t)} &, \text{else}
    \end{cases}\,\,.
\end{align}

Now, consider $\mathcal{L}_{\text{NCA++}}^{\text{L2}}$, i.e., $d(\rho, \zeta) = \lVert \nu_p - \nu_z \rVert^2 = \kappa_p^2 + \kappa_z^2 - 2 \kappa_p \kappa_z s(\mu_p, \mu_z)$, following from the law of cosines. Here, $\frac{\delta}{\delta s} d(\nu_p, \nu_z) = - 2 \kappa_p \kappa_z$, which we can again plug into Equations~\ref{form:diff_loss} and \ref{form:diff_lse} and obtain:
\begin{align}
    \frac{\delta}{\delta s} \mathcal{L}_{\text{NCA++}}^{\text{L2}} &= \begin{cases}
    -\frac{2 \kappa_p \kappa_z}{t} + \frac{2 \kappa_p \kappa_z}{t} \frac{\exp(-d(\rho, \zeta)/t)}{\sum_{c=1}^C \exp(-d(\rho_c, \zeta)/t)} &, \text{if $p=p^*$} \\
    \frac{2 \kappa_p \kappa_z}{t} \frac{\exp(-d(\rho, \zeta)/t)}{\sum_{c=1}^C \exp(-d(\rho_c, \zeta)/t)} &, \text{else}
    \end{cases} \\
    &= \begin{cases}
    -\frac{2 \kappa_p \kappa_z}{t} + \frac{2 \kappa_p \kappa_z}{t} \frac{\exp((\kappa_p^2 + 2 \kappa_p \kappa_z s(\mu_p, \mu_z))/t)}{\sum_{c=1}^C \exp((\kappa_{p_c}^2 + 2 \kappa_p \kappa_z s(\mu_{p_c}, \mu_z)) / t)} &, \text{if $p=p^*$} \\
    \frac{2 \kappa_p \kappa_z}{t} \frac{\exp((\kappa_p^2 + 2 \kappa_p \kappa_z s(\mu_p, \mu_z))/t)}{\sum_{c=1}^C \exp((\kappa_{p_c}^2 + 2 \kappa_{p_c} \kappa_z s(\mu_{p_c}, \mu_z))/t)} &, \text{else}
    \end{cases} \,\,.
\end{align}

\section{Summary of Loss Calculation}\label{sec:alg_summary}

Algorithm~\ref{alg:elnivmf} sketches how \textbf{EL-nivMF} is implemented practically. As discussed, the parameters of the proxies are learnable parameters, whereas the vMF distributions of points are predicted by an encoder. Thus, the module in Algorithm~\ref{alg:elnivmf} can be plugged on-top of an encoder and trained jointly. Since test-time retrieval only requires access to the image-embeddings, the module can be discarded after training.

\begin{algorithm}
\DontPrintSemicolon
\caption{Module to compute \textbf{EL-nivMF} loss}\label{alg:elnivmf}
\SetKwFunction{FMain}{initialize}
    \SetKwProg{Fn}{Function}{:}{}
    \Fn{\FMain{$C$: num proxies, $M$: dimensions, $N$: num samples}}{
        $\mu_\rho \gets \text{learnable tensor} \in [C,M]$\;
        $\kappa_\rho \gets \text{learnable tensor} \in [C,M]$\;
        $t \gets \text{learnable parameter} \in [1]$\;
        Save $C, M, N$\;
    }
    
\SetKwFunction{Floss}{loss}
\Fn{\Floss{$z$: image embedding $\in [1, M]$, $c^*$: ground-truth proxy index}}{
        \texttt{samples} $\gets$ empty matrix $\in [N, D]$\;
        \For{$n=1,\dotsc,N$}{
            \texttt{samples}$[n,:] \sim \text{vMF}\left(\mu = \frac{z}{\lVert z \rVert}, \kappa = \lVert z \rVert \right)$\;
        }
        \texttt{sim\_to\_proxy} $\gets$ empty vector $\in [C]$\;
        \For{$c=1, \dotsc, C$}{
            \texttt{logls} $\gets$ empty vector $\in [N]$\;
            \For{$n=1,\dotsc,N$}{
                \texttt{logls}$[n]$ $= \log ( \texttt{nivmf\_likelihood}(z, \mu = \mu_\rho[c,:], K = \text{diag}(\kappa_\rho[c,:]))$\;
            }
            \texttt{sim\_to\_proxy}$[c] \gets \texttt{logsumexp}(\texttt{logls} / t)$\;
        }
        \texttt{logloss} $\gets -\texttt{sim\_to\_proxy}[c^*] + \texttt{logsumexp}(\texttt{sim\_to\_proxy})$\;
        \Return{\texttt{logloss}}
    }
\end{algorithm}

\section{Experimental Details}\label{supp:exp_details}
As already noted in \S\ref{sec:distrdist}, we generally utilize $N\approx 10$ for our Monte-Carlo estimation of the PPK kernel (Eq.~\ref{form:elnivmf}), but switch to $N = 5$ for hyperparameter searches and $N = 20$ for our ablation experiments, as within this range, we found performance to be similar.

\section{Experimental Details Ablation Study}

To reduce any influences of covariates, we seek to keep experimental settings in the ablation study in \S\ref{sec:ablation_metrics} constant across all benchmarked metrics. Hence, we fixed all hyperparameters as in the previous experiment, and tuned the following hyperparameters for each approach on validation data:
\begin{align}
    t &\in \{ 1, 1/32, 1/256\} \\
    \kappa_p &\in \{10, 50, 200\} \text{ (for \textbf{ni-vMF}, this is for each dimension)} \,.
\end{align}
Across all metrics, we used the dimensionality $M=512$, a batchsize of $106$, and $150$ epochs on CARS and $50$ on CUB. To reduce the initialization noise, we initiated each hyperparameter-tuning experiment $3$ times with random seeds, then calculated the median of the maximum $R@1$ performance on the validation set, and ran the best hyperparameter settings with $5$ seeds. 




\section{$L_2$ Distance as Retrieval Metric}
\label{sec:l2retrieval}

\begin{table}[h]
\setlength{\tabcolsep}{6pt}
    \centering
    \caption{R@1 of the same trained models from Figure~\ref{fig:abl}, but using the  euclidean instead of the cosine distance for retrieval.}
    \label{tab:ablation_quant}
    \vskip\baselineskip
    \begin{tabular}{lcc}
        \toprule
        Method & CUB & CARS\\
        \midrule
        $d_{\text{L2}}$ & $61.89 \pm 0.36$ & $76.61 \pm 0.17$\\
        $d_{\text{Cos}}$ & $62.01 \pm 0.35$ & $76.94 \pm 0.49$\\
        $d_{\text{nivMF}}$ & $63.74 \pm 0.18$ & $78.62 \pm 0.41$ \\
        \midrule
        $d_{\text{B-vMF}}$ & $62.29 \pm 0.34$ & $79.69 \pm 0.15$ \\
        $d_{\text{EL-vMF}}$ & $62.49 \pm 0.56$ & $80.17 \pm 0.24$ \\
        $d_{\text{KL-vMF}}$ & $61.68 \pm 0.36$ & $76.65 \pm 0.20$ \\
        $d_{\text{EL-nivMF}}$ & $63.69 \pm 0.56$ & $76.37 \pm 5.32$\\
        \bottomrule
    \end{tabular}
\end{table}

\section{Qualitative Impact on Image Norms}
\label{sec:norm_hist}
To understand in more detail the difference in learned and assigned image norms produced when training with $d_\text{EL-nivMF}$, we compare the distribution of image norms between those belonging to originally correctly and incorrectly classified samples (initial separation done using a standard baseline DML model operating on $d_\text{cos}$) for CUB \& CARS, respectively. 
\begin{figure}[h!]
    \vspace{-10pt}
    \centering
    \includegraphics[width=0.95\textwidth]{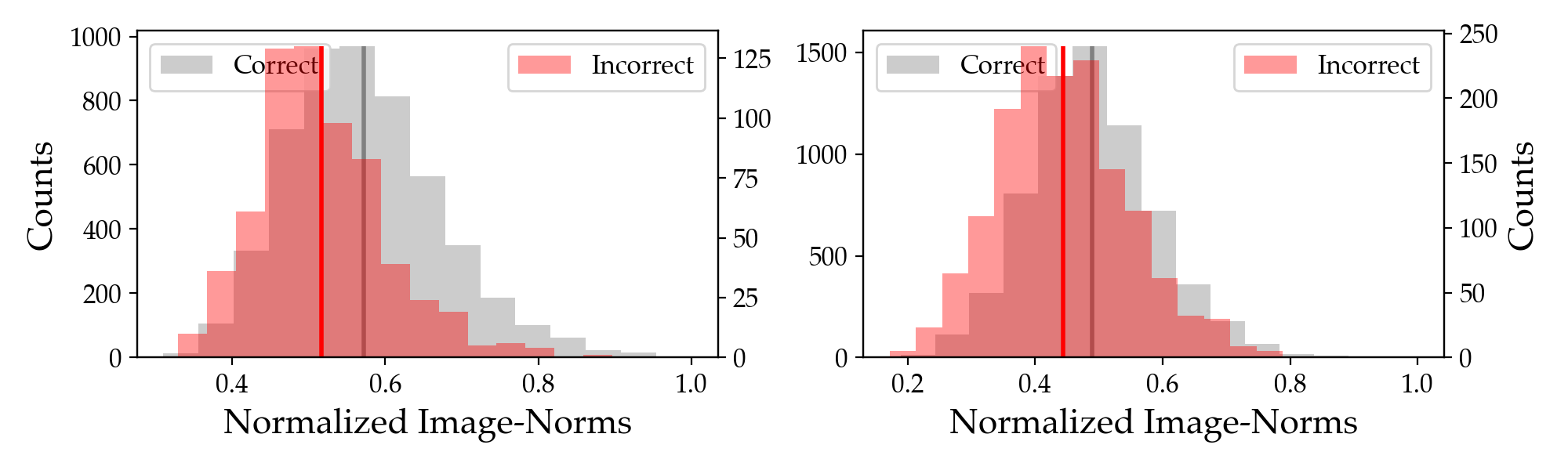}
    \caption{Norms of prev. \textcolor{gray}{correct}/\textcolor{red}{incorrect} pred. on CUB/CARS.}
    \label{fig:norms}
\end{figure}
Results are shown in Fig.~\ref{fig:norms}, which reveal that correct classifications on average have higher norms while miss-classifications are more often attributed to lower norms. This aligns well with the underlying motivation assigning low norms to ambiguous images (compare to e.g. Sec.\ref{sec:norms}).

\section{Non-isotropic Proxies Encourage Diverse Representations}
\label{sec:furtherablations}
Finally, we qualitatively investigate the metric representation spanned by metric learners trained using $d_\text{EL-nivMF}$.
To do so, we follow both \cite{roth2020revisiting} and look at the feature diversity, as well as evaluating the cluster diversity to see whether encouraging unique class-proxy distributions helps in learning a more diverse class-specific encoding.
For the former, we follow \cite{roth2020revisiting} and evaluate the uniformity of the sorted spectral value distribution of all training image embeddings to measure the number of significant directions of variances in feature space.
The latter is simply computed as the variance (i.e. diversity) of intraclass distances for each class-cluster.
For both cases, we specifically care about relative changes compared to models trained without probabilistic treatment (i.e. using $d_\text{cos}$) as well as changes going from an isotropic ($d_\text{EL-vmf}$) to a non-isotropic setup ($d_\text{EL-nivMF}$).
\begin{table}[h!]
    \centering
  \resizebox{0.7\textwidth}{!}{
    \begin{tabular}{cccc}
    \toprule
        Dataset & Metric & $d_\text{cos} \rightarrow d_\text{EL-vMF}$ & $d_\text{cos} \rightarrow d_\text{EL-nivMF}$ \\
        \midrule
        \multirow{2}{*}{CARS} & Cluster-Div.$\uparrow$  & $+24\%$ & $+31\%$ \\
        & Feat.-Div. $\uparrow$ & $+13\%$  & $+14\%$\\
        \midrule
        \multirow{2}{*}{CUB} & Cluster-Div. $\uparrow$ & $+11\%$ & $+25\%$ \\
        & Feat.-Div. $\uparrow$ & $+6\%$  & $+8\%$\\
    \bottomrule
    \end{tabular}}
    \caption{Metrics on how \textbf{EL-nivMF} structures the embeddings.}
    \label{tab:metric_changes}
\end{table}
Results are summarized in Tab.~\ref{tab:metric_changes}, showcasing a consistent improvement in both feature and cluster diversity when incorporating both a probabilistic treatment and a non-isotropic encoding of proxy distributions. This provides further heuristic evidence linking the usage of $d_\text{EL-nivMF}$ to a better capture of the semantic class variability as well as an improved incorporation of a more diverse feature set, shown to facilitate generalisation \cite{roth2020revisiting,milbich2020diva}.

\newpage
\section{Further Qualitative Embedding Norm Studies}
\label{sec:extra_norm}

\begin{figure*}[h]
    \centering
    \includegraphics[width=0.42\linewidth]{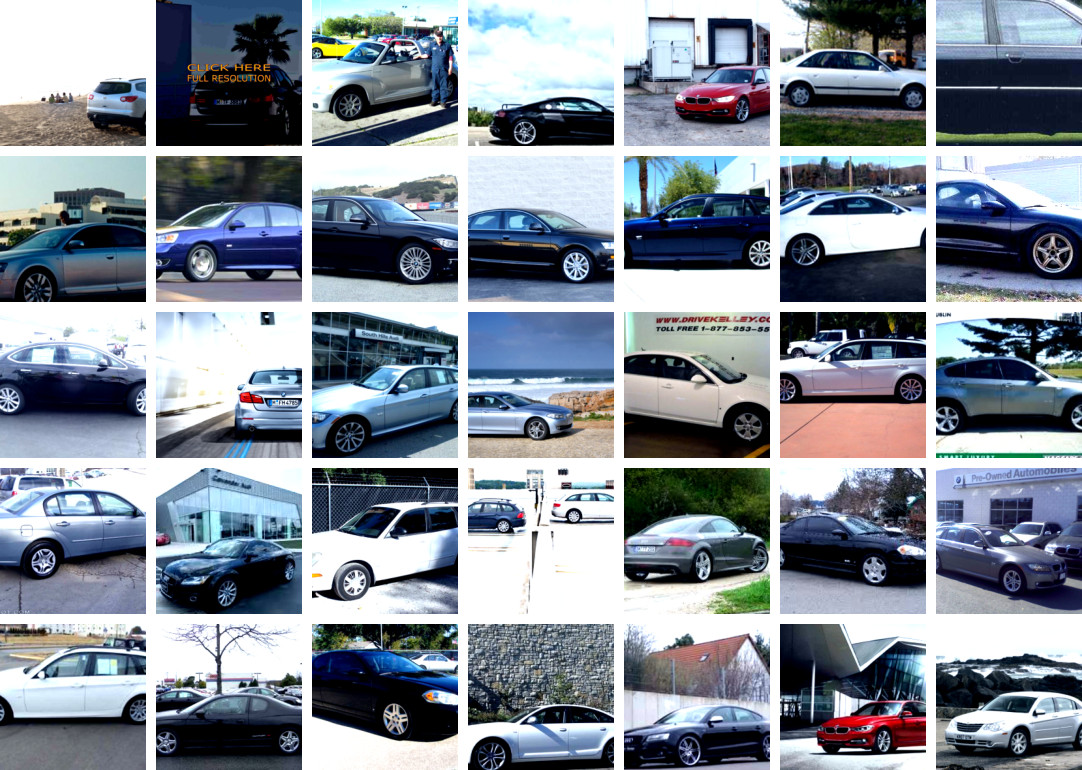}
    \begin{tikzpicture}
        \node (n1) {$\dots$}; 
        \node (n2) at ([yshift=0.8cm]n1) {$\dots$}; 
        \node (n3) at ([yshift=1.6cm]n1) {$\dots$}; 
        \node (n4) at ([yshift=2.4cm]n1) {$\dots$}; 
        \node (n5) at ([yshift=3.2cm]n1) {$\dots$}; 
    \end{tikzpicture}
    \includegraphics[width=0.42\linewidth]{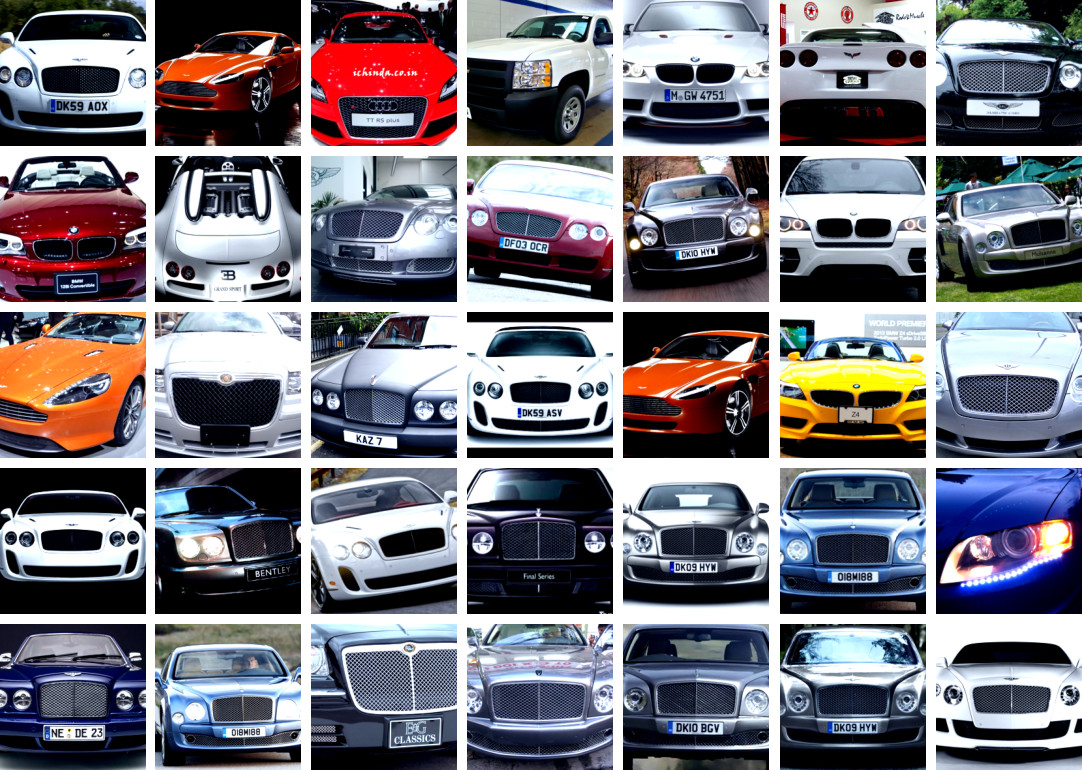}
    
    \begin{tikzpicture}
        \node (n1) {};
        \node (n3) [below of= n1, xshift=1cm, yshift=0.7cm]{lowest norm};
        \node (n2) at ([xshift=11.38cm]n1) {};
        \node (n4) [below of= n2, xshift=-1cm, yshift=0.7cm]{highest norm};
        \draw [-stealth] (n1) -- (n2);
    \end{tikzpicture}
    \caption{CARS train images with lowest (left) to highest (right) embedding norms on a $M=512$ dimensional ResNet-50 backend.}
    \label{fig:qual_norm_global}
\vspace{-10pt}
\end{figure*}
\begin{figure*}[h]
    \centering
    \includegraphics[width=0.4\linewidth]{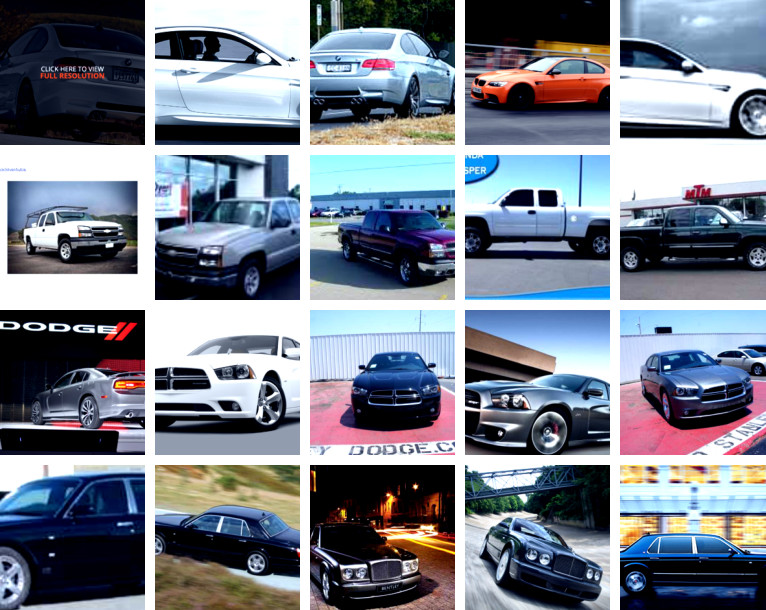}
    \begin{tikzpicture}
        \node (n1) {$\dots$}; 
        \node (n2) at ([yshift=1cm]n1) {$\dots$}; 
        \node (n3) at ([yshift=2cm]n1) {$\dots$}; 
        \node (n4) at ([yshift=3cm]n1) {$\dots$}; 
    \end{tikzpicture}
    \includegraphics[width=0.4\linewidth]{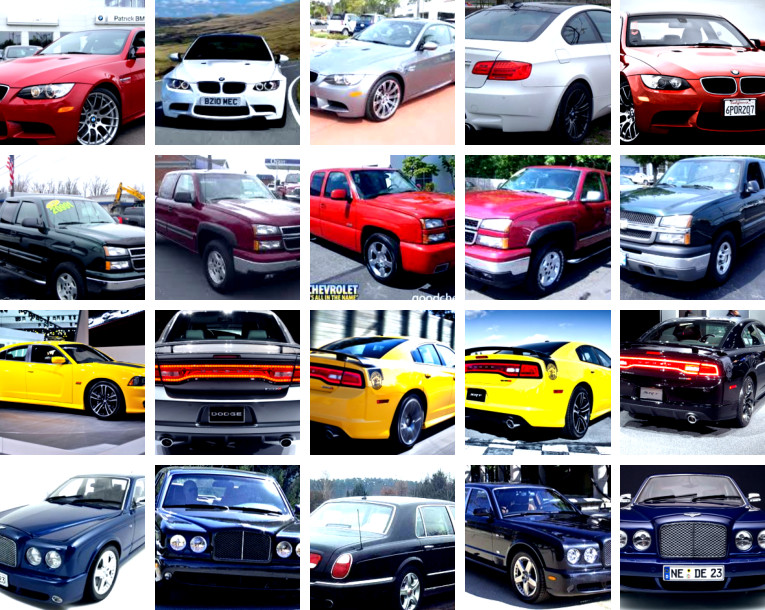}
    
    \begin{tikzpicture}
        \node (n1) {};
        \node (n3) [below of= n1, xshift=1cm, yshift=0.7cm]{lowest norm};
        \node (n2) at ([xshift=10.87cm]n1) {};
        \node (n4) [below of= n2, xshift=-1cm, yshift=0.7cm]{highest norm};
        \draw [-stealth] (n1) -- (n2);
    \end{tikzpicture}
    \caption{Images for four randomly chosen classes (rows) of the CARS training set, ordered by their norm from lowest (left) to highest (right). Obtained from the $d_{\text{EL-vMF}}$ model on a ResNet-50, where the norms of image embeddings range from $70.58$ to $140.09$ whereas the proxy norms are between $45.95$ to $79.98$.}
    \label{fig:qual_norm}
\end{figure*}


%
%

\end{document}